\pdfoutput=1
\documentclass[11pt]{article}

\usepackage[hyperref]{acl}
\usepackage{times}
\usepackage{latexsym}
\usepackage{enumitem}
\usepackage{graphicx}
\usepackage{booktabs}
\usepackage{multirow}
\usepackage{hyperref}
\usepackage{xcolor}
\usepackage{amssymb}
\usepackage{xspace}
\usepackage{comment}
\usepackage{etoolbox,caption,subcaption}
\usepackage{microtype}

\definecolor{situation_color}{HTML}{1B9E77}
\definecolor{joke_color}{HTML}{D95F02}
\definecolor{explanation_color}{HTML}{7570B3}
\definecolor{error_color}{HTML}{FF0000}
\definecolor{junglegreen}{rgb}{0.16, 0.67, 0.53}
\definecolor{salmon}{HTML}{F69289}  %
  
\providetoggle{showhidden}
\toggletrue{showhidden} %

\providetoggle{pubstamp}
\togglefalse{pubstamp}
\providecommand{\pubstampit}{
\iftoggle{pubstamp}{
\vspace*{-.1cm}\textcolor{red}{\fbox{{\it Publication venue:} ACL 2023\xspace}}} %
{}
}

\newcommand{\matching}{matching\xspace}
\newcommand{\Matching}{Matching\xspace}
\newcommand{\ranking}{quality ranking\xspace}
\newcommand{\Ranking}{Quality ranking\xspace}
\newcommand{\explanation}{explanation\xspace}
\newcommand{\Explanation}{Explanation\xspace}

\newcommand{\ncontests}{704\xspace}
\mathchardef\mhyphen="2D
\newcommand{\nodata}{-}

\makeatletter
\newenvironment{chapquote}[2][2em]
  {\setlength{\@tempdima}{#1}%
   \def\chapquote@author{#2}%
   \parshape 1 \@tempdima \dimexpr\linewidth-2\@tempdima\relax%
   \itshape}
  {\par\normalfont\hfill--\ \chapquote@author\hspace*{\@tempdima}\par\bigskip}
\makeatother

\renewcommand{\sectionautorefname}{\S\kern-3pt}  
\renewcommand{\subsectionautorefname}{\S\kern-3pt}

\title{
Do Androids Laugh at Electric Sheep?
Humor ``Understanding" Benchmarks from
The New Yorker Caption Contest
}

\author{
    \begin{tabular}{c}
    Jack Hessel$^\dagger$ \quad Ana Marasovi\'{c}$^{\diamond}$\quad Jena D. Hwang$^\dagger$ \quad Lillian Lee$^\circ$ \vspace{.5mm} \\
    Jeff Da$^\ddag$\quad Rowan Zellers$^{\bullet}$\quad Robert Mankoff$^\blacktriangle$\quad Yejin Choi$^{\dagger\ddag}$
    \end{tabular}
    \\ \vspace{.5mm}
    \small
    \begin{tabular}{c}
    $^\dagger$ The Allen Institute for AI \quad $^\diamond$ University of Utah \quad
    $^\circ$ Cornell University \quad $^\bullet$OpenAI \\
    $^\ddag$ %
    University of Washington \quad
    $^\blacktriangle$ 
    Air Mail and Cartoon Collections\\
    \end{tabular}
    \\ \vspace{.5mm}
    \small
    \begin{tabular}{c}
    \texttt{jackh@allenai.org} \quad \texttt{ana.marasovic@utah.edu} \quad \texttt{jenah@allenai.org} \quad \texttt{llee@cs.cornell.edu}  \\
    \texttt{\{jzda,rowanz\}@cs.washington.edu} \quad \texttt{bob@bobmankoff.com} \quad \texttt{yejin@cs.washington.edu}
    \end{tabular}
    \vspace{2mm} \\
    \pubstampit
}

\date{}

\begin{document}
\maketitle
\begin{abstract}

Large neural networks can now generate jokes, but do they really ``understand''
humor? 
We challenge AI models with three tasks derived from the New Yorker Cartoon Caption Contest: matching a joke to a cartoon, identifying a winning caption, and explaining why a winning caption is funny. 
These tasks 
encapsulate progressively more sophisticated aspects of ``understanding'' a cartoon;
key elements are the complex, often surprising relationships between images and captions
and the frequent inclusion of indirect and playful allusions to human experience and culture. 
We investigate %
both multimodal and language-only models:
the former are challenged with the cartoon images directly, while the latter are given multifaceted descriptions of the visual scene to simulate human-level visual understanding.
We find that both types of %
models struggle at all three tasks. %
For example, our best multimodal models fall 30 accuracy points behind human performance on the matching task, and, even when provided ground-truth visual scene descriptors, human-authored explanations are preferred head-to-head over the best machine-authored ones (few-shot GPT-4) in more than 2/3 of cases.
We 
release 
models, code, leaderboard, and corpus, which includes newly-gathered annotations describing the image's locations/entities, what's unusual in the scene, and an explanation of the joke.

\end{abstract}

\section{Introduction}

\begin{chapquote}{\citet{dissectfrog}}
Humor can be dissected, as a frog can, but the thing dies in the process and the innards are discouraging to any but the pure scientific mind.
\end{chapquote}
\nocite{dissectfrog}

\begin{figure}
    \centering
    \includegraphics[width=\linewidth]{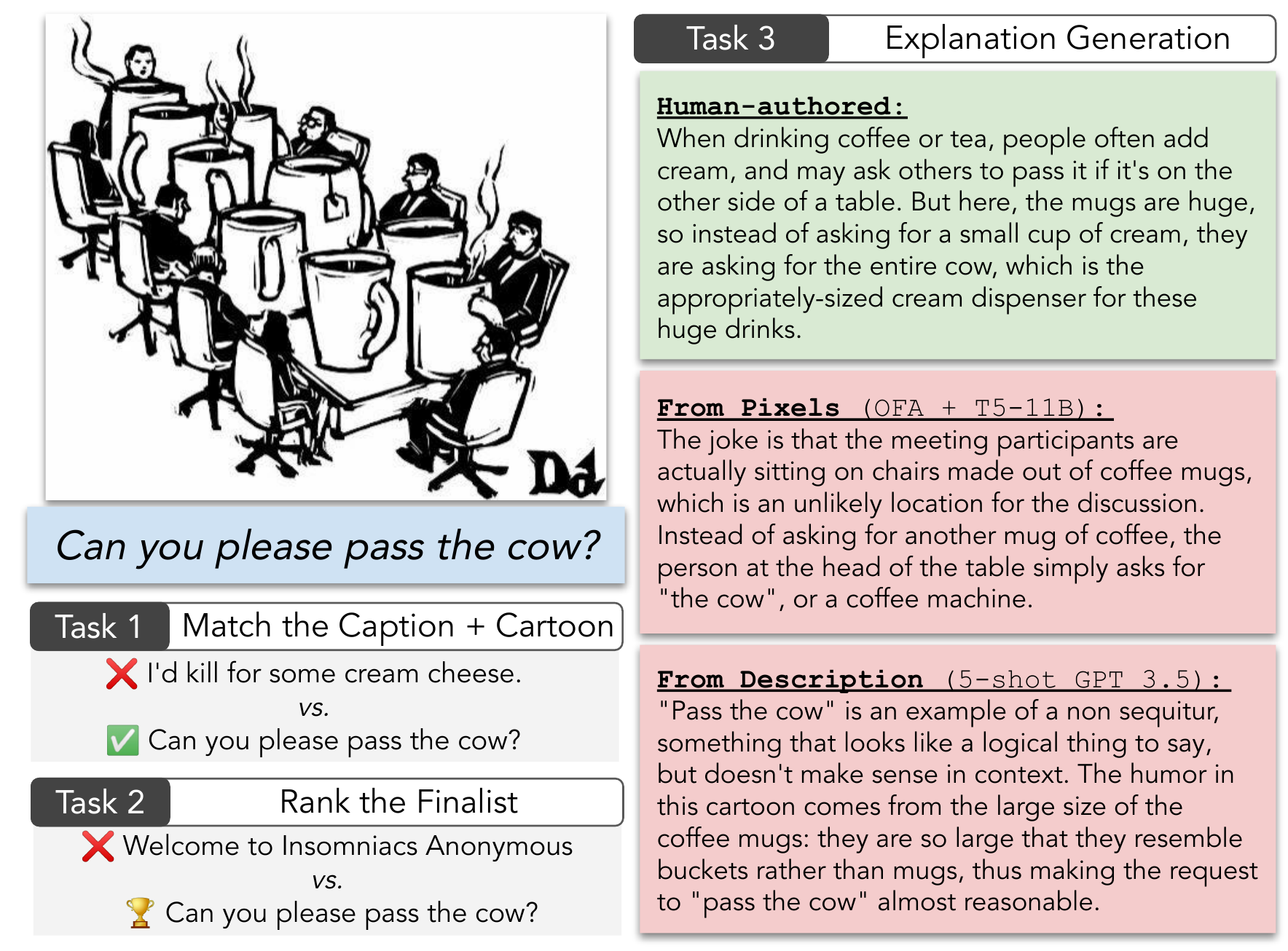}
    \caption{%
    We formulate three tasks using over a decade of New Yorker caption contests: models must 1) {recognize} a caption written about a cartoon (vs.~options that were not); 2) {evaluate} that caption's ``quality'' by scoring it more highly than a non-finalist/non-winner from the same contest; and 3) explain why the joke is funny.
    (Cartoon by Drew Dernavich, winning caption by Bennett Ellenbogen).
    }
    \label{fig:fig1}
\end{figure}

\vspace{-.3cm}

\begin{figure*}
\centering
\includegraphics[width=.95\linewidth]{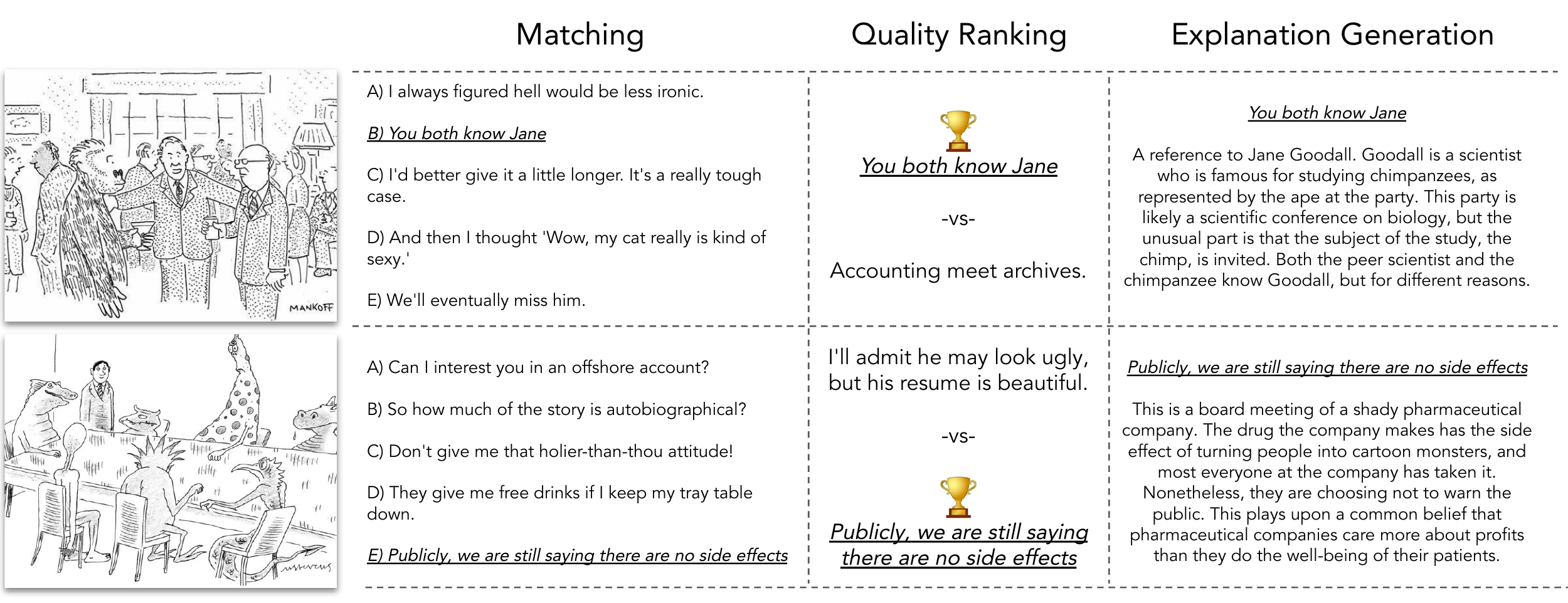}
\caption{Instances of our three tasks. 
\underline{\emph{\Matching}} requires models to select the finalist caption for the given cartoon from among distractors that were finalists, but for {\em other} contests. \underline{\emph{\Ranking}} requires models to differentiate a finalist from a non-finalist, both written for the given cartoon. \underline{\emph{\Explanation}} requires models to generate free-text explanations of how a high-quality caption %
relates to the cartoon. Cartoons by Robert Mankoff and Mick Stevens.}
\label{fig:task_setup_in_detail}
\end{figure*}

Each week, {\em The New Yorker}
publishes a un-captioned cartoon image,
inviting readers
to submit their funniest English-language caption for it. Editors choose three finalists from sometimes thousands of submissions; then,
readers vote to pick the final winner.
We develop a suite of three progressively harder tasks built around this contest to test how well AI models ``understand'' humor across vision and language: 1) matching jokes to cartoons, 2) identifying a winning caption, and 3) generating an explanation of why an image/caption combination is funny.

These tasks are difficult because 
the connection between a winning caption and image can be quite subtle, and the 
caption can make
playful allusions to human experience, culture, and imagination.
Consider the image and winning caption ``Can you please pass the cow?'' in \autoref{fig:fig1}.
Unlike literal image captions such as in MSCOCO \cite{lin2014microsoft}, here, the caption's relation to the image is indirect:\footnote{The (relatable) experience of ``not getting'' a New Yorker cartoon often results from inability to identify the image/text relationship.} the size of the mugs must first be recognized as unusual, and then, the caption invokes an association between a large mug and a large amount of cream/milk --- perhaps a whole cow's worth. 
Further, matching a caption to an image is not sufficient: non-finalist
entries (e.g., ``...Insomniacs Anonymous'' in \autoref{fig:fig1}) also match the image, but something else makes one seem funnier than the other.
Finally, even if a model can accurately identify winning submissions, we would like it to also be able to explain {\em why} a particular highly rated/relevant caption is funny. %

We cover our three tasks in two settings: in the \emph{from pixels} setting, models are given access only to the cartoon images at test time, and must perform computer vision;
in the \emph{from description} setting, we allow models access to a newly-collected, human-authored corpus of cartoon descriptions, %
thus simulating access to a
human-level
computer-vision system --- or, alternately, facilitating benchmarking of models
that don't have a built-in image-processing component.
The annotations we collect and release are rich and multifaceted: they describe the image 
overall and its locations and entities,
what's unusual about the image, and an explanation of the joke. We view this effort as a significant contribution of our work.

Our results reveal 
a gap
between AI and human-level humor ``understanding.''
In the \emph{from pixels} setting, our best multimodal model (fine-tuned CLIP ViT-L/14 \cite{radford2021learning}) achieves 62\% accuracy on a 5-way multiple choice task, but humans achieve 94\% in the same setting. Even with
significant manual annotation of the cartoons %
in the \emph{from description} setting (and
despite significant improvements in language modeling performance
since this work's submission\footnote{GPT-3 \cite{brown2020language} was the most performant in Jan. 2023 when this work was submitted, but we have since updated our results.})
large language models still fall short:
human explanations are still preferred in more than two-thirds
of cases compared to our best explanation model, 5-shot GPT-4.

We release our challenging NLP/vision benchmarks,\footnote{Our data may contain offensive jokes. We manually removed a handful of cases we observed to target specific protected classes.
We do not endorse the jokes in the corpus, but rather, view them as interesting objects of study.} annotations, models, leaderboard, and code at \url{https://capcon.dev/}. Beyond AI research, we also hope that our work will 
spur progress in human-AI collaboration tools for cartoonists, contest entrants, and beyond (see~\autoref{sec:sec_with_human_AI_collaboration} for AI-generated captions).

\section{Datasets and Task Setups}

Our corpus compiles 14 years of weekly New Yorker caption contests. 
Each contest consists of:
(1) a captionless cartoon; (2) that week's entries; (3) the three finalists, selected by New Yorker editors; and (4) for some contests, quality estimates for each submission collected via crowdsourcing.\footnote{We regret that The New Yorker does not
currently have an alliterative-paragraph contest.} %

\label{sec:sec_with_original_corpus_details}
The corpus was constructed from two sources. The first is
\newcite{newyorkernextmldataset}, from which
we obtain roughly 250 contests (mean/median 6.1K/5.7K unique captions per contest; 1.5M total), starting from \#508.\footnote{
We manually corrected \href{https://github.com/nextml/caption-contest-data-api/issues/38}{some errors} in the corpus.
}
Crowd ratings in this corpus
are gathered via the \href{http://nextml.org/}{NEXT platform} \cite{jamieson2015next,NIPS2017_c02f9de3}, where
readers rate captions as ``funny", ``somewhat funny", or ``unfunny"; we use the per-caption mean. %
There are over 114M ratings total (mean/median of 445K/471K per contest). %
We also sample three additional top captions that aren't editorial picks to serve as additional ``finalists."

The second corpus, due to \citet{shahaf2015inside,radev-etal-2016-humor} and derived from contests \#1--\#507,
includes 2M unique captions (mean/median 5.2K/5.0K per contest), but no crowd ratings. 
We remove by hand 55 contests 
whose images' resolutions are too low, and 
identify 80 low resolution (but usable) cases, taking special care when annotating this set (\autoref{sec:sec_with_manual_annotation}).

\subsection{Task Setups}

\begin{table}[t]
    \resizebox{.99\linewidth}{!}{
    \centering
    \begin{tabular}{lc}
    \toprule
        \# Train/val/test \Matching & 1.6K / 538 / 538 \\
        \# Train/val/test \Ranking &
         1.6K / 523 / 523 \\
        \# Train/val/test \Explanation & 391 / 130 / 130\\
    \bottomrule
    \end{tabular}
    }
    \caption{Basic size statistics for our three tasks. We extend %
    \newcite{shahaf2015inside,radev-etal-2016-humor,newyorkernextmldataset} by (a) proposing \matching, \ranking, and \explanation tasks; (b) providing new, dense annotations for each cartoon (see~\autoref{fig:annotations}); (c) authoring a set of 651 joke explanations.}
    \label{tab:basic_statistics}
\end{table}

We pose three tasks. %
\Matching and \explanation are novel, whereas \ranking extends the formulations introduced in \newcite{shahaf2015inside,radev-etal-2016-humor}.

\paragraph{\Matching.}\emph{Can a model recognize when a caption is appropriate for a given cartoon?} Five choices are given, only one of which truly corresponds. For the example in \autoref{fig:fig1}, we supply the following possibilities: %
\begin{enumerate}[label=(\alph*),leftmargin=*,topsep=0pt,itemsep=-1ex,partopsep=1ex,parsep=1ex]
    \item \emph{O.K. I'm at the window. To the right? Your right or my right?}
    \item \emph{I'd kill for some cream cheese.}
    \item \emph{Bob just came directly from work.}
    \item \textbf{Can you please pass the cow?}
    \item \emph{They only allow one carry-on.}
\end{enumerate}
The correct caption is a finalist for the cartoon. %
Negative choices are randomly selected finalists from other contests, 
and as a result, 
are 
great captions for some {\em other} contest's image.\footnote{Distractors are balanced across instances so that %
a model that only examines the answer choices cannot achieve better than chance accuracy.} 
In some cases, matching depicted objects to their textual references may suffice, but in other cases, the relationship is more indirect. For example, \autoref{fig:task_setup_in_detail} (top) contains a subtle reference to Jane Goodall, thus requiring external knowledge;  
\autoref{fig:task_setup_in_detail} (bottom) relies on a stereotype of pharmaceutical companies being untrustworthy, hence requiring reasoning
beyond the literal text.

\paragraph{\Ranking.}\emph{Can a model identify highly rated captions?}
For each finalist, we sample for comparison a caption that was \emph{not} selected as a finalist, and ask models to identify which one (the real one or the distractor) was rated as higher quality.
As preprocessing, 
we run one round of text-only filtering to discard submissions that are easily identifiable as low quality, and also perform semantic deduplication; more details in \autoref{sec:task_construction_details}.
Here is the end result for \autoref{fig:fig1}:

\begin{figure*}
\centering
\includegraphics[width=.95\linewidth]{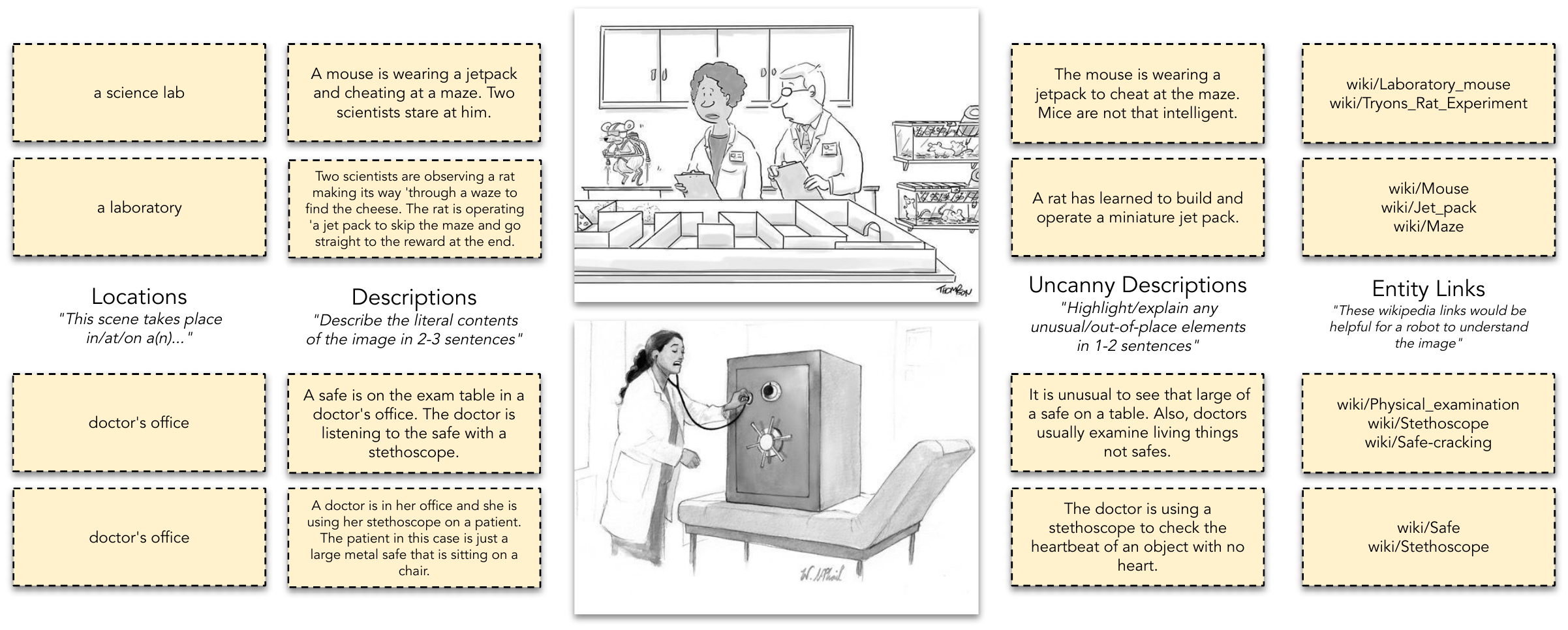}
\caption{For each of \ncontests cartoons, we gather several types of annotations from human participants, including locations, descriptions, descriptions of uncanniness, and relevant entities in the form of English Wikipedia links. Annotations shown are true random samples from the corpus. Cartoons by Mark Thompson and Will McPhail.
}
\label{fig:annotations}
\end{figure*}

\begin{enumerate}[label=(\alph*),leftmargin=*,topsep=0pt,itemsep=-1ex,partopsep=1ex,parsep=1ex]
    \item \textbf{Can you please pass the cow?}
    \item Welcome to Insomniacs Anonymous.
\end{enumerate}
Which caption a particular individual prefers can be a matter of personal taste;
but there is a general preference among our human annotators for the true finalist  (see \autoref{sec:sec_with_clf_experiments}).

\paragraph{\Explanation.}\emph{Can a model generate 
as good an explanation as a human
for why a caption-and-image combination is funny?} 
Free-form explanations of why captions are funny/appropriate for their corresponding image were written by an author of this paper.%
\footnote{Several attempts to solicit explanations from crowdworkers were not satisfactory; similarly unsuccessful were prompting experiments with GPT-3 inspired by \newcite{wiegreffe2022reframing,marasovic2021few} --- too few of the sampled explanations were correct to bootstrap a corpus.}
The rough annotation guidance was: ``In a few sentences, explain the joke as if to a friend who doesn't `get it' yet."  %
Starting from a random finalist for each contest, after filtering out cases where the author did not understand the joke, a corpus of 651 human-created joke explanations to serve as comparison points was formed (mean/median 60/59 words, 39.3K total).
We consider a model to succeed at this task if human judges, presented with (unlabeled) 
pairs of author/machine-generated explanations, do not show a preference for the author-generated ones.

\paragraph{Evaluation metrics.} For \matching and \ranking, we evaluate using accuracy. %
For \ranking, we report {\em NYAcc} --- the average accuracy over instances where the finalist was an official New Yorker finalist ---  and {\em CrowdAcc}, where the ``finalist'' caption was selected by the crowd as high quality. These two measures allow us to account for different
audience tastes.
For \explanation, we conduct pairwise human evaluations to test several hypotheses detailed in \S\ref{sec:sec_with_pairwise_comparisons}. To complement these human evaluations, we also report in \autoref{sec:sec_on_auto_eval} automatic metrics that take into account the human-written reference: (a) BLEU-4 \cite{papineni2002bleu} using \newcite{post2018call}+ROUGE-L \cite{lin2004rouge}; and (b) word-level perplexity. %

\paragraph{From Pixels + From Description.}
We consider two experimental settings. In \textbf{From Pixels (FP)}, a vision+language model undertakes image processing, %
i.e., at test time, the only contest information available is the image itself.
In the second setting, which we call \textbf{From Description (FD)}, we factor out visual processing by providing the model with human written annotations, described in \S\ref{sec:sec_with_manual_annotation}. 
FD models thus simulate access to a
human-level
computer-vision system.

\subsection{Annotation of cartoons.}

\label{sec:sec_with_manual_annotation}

We collect
several types of annotations about the \ncontests cartoons; these either serve as input to models in the \emph{from description} setting, or as additional information available only at training time in the \emph{from pixels} setting. For each cartoon, we gather:

\begin{enumerate}[label=(\roman*),leftmargin=*,topsep=0pt,itemsep=-1ex,partopsep=1ex,parsep=1ex]
    \item A phrase describing the setting of the scene, e.g., ``an office" or ``the park" (2 per cartoon)
    \item A literal 1-3 sentence description of the scene (3 per cartoon)
    \item A 1-3 sentence description or explanation of what makes the scene unusual (3 per cartoon)
    \item 2-3 English Wikipedia links that an annotator identified as relevant,
    to serve as a proxy for world knowledge (2 per cartoon) %
\end{enumerate}

A random sample of annotations is shown in~\autoref{fig:annotations}. %
We used Amazon Mechanical Turk, and paid crowdworkers a minimum of \$15/hr. %
Low-resolution images involved special treatment:  1) we offered additional pay to crowdworkers; and 2) at least one of the annotations is conducted by an author of this work using the same HIT interface. %
Details including qualification rounds, screenshots of the HITs, etc. are given in \autoref{sec:sec_with_crowdworking_details}.

\section{Experiments}
\label{sec:sec_with_clf_experiments}

We split the \ncontests cartoons into 5 cross-validation splits such that entire contests are held out at test time. Task construction details are in \autoref{sec:task_construction_details}; modeling details (e.g., hyperparameter sweeps, task formatting) are in~\autoref{sec:sec_with_additional_experimental_details}.

\subsection*{From Pixels (FP) Models}

We explore two vision+language models.

\paragraph{CLIP.} We fine-tune CLIP \texttt{ViT-L/14@366px} \cite{radford2021learning} (428M parameters), which consists of a text Transformer \cite{vaswani2017attention} and a vision Transformer \cite{dosovitskiy2020image} pretrained to align %
images/captions in the WebImageText corpus
(400M pairs). %
For multiple choice, we %
use InfoNCE \cite{oord2018representation} to encourage the cosine similarity of the cartoon/correct answer to be higher than the incorrect ones. %
For zero-shot classification, we use the prompt \texttt{a new yorker cartoon with winning caption}. CLIP isn't generative, so we can't use it for \explanation.

\newcommand{\zshotres}{\hspace*{.4in}\rotatebox[origin=c]{180}{$\Lsh$}\xspace}
\newcommand{\fiveshotres}{\zshotres}
\begin{table*}[t]
        \begin{minipage}[t]{0.6\textwidth}
        \resizebox{\textwidth}{!}{          
            \begin{tabular}{clccc}
              & & \multicolumn{1}{c}{Matching} & \multicolumn{2}{c}{Quality Ranking}
              \\
              \cmidrule(lr){3-3}\cmidrule(lr){4-5}
              & & Accuracy ($\uparrow$) & CrowdAcc ($\uparrow$) & NYAcc ($\uparrow$)\\
              \midrule
               & Random & 20.0 & 50.0 & 50.0  \\
              & Caption Only (T5-11B) & 19.4 & 59.4 & 64.5  \\
              \midrule
              \parbox[t]{1.5mm}{\multirow{4}{*}{\rotatebox[origin=c]{90}{FP}}}
               & CLIP ViT-L/14@336px (finetuned) & \underline{62.3} & 57.0 & \underline{66.9} \\
               & \quad \rotatebox[origin=c]{180}{$\Lsh$} Zero-shot & \zshotres 56.6 & \zshotres 55.8 & \zshotres 56.8\\
               & OFA-Huge $\rightarrow$ T5-Large & 45.2 & 59.1 & 64.3 \\
               & OFA-Huge $\rightarrow$ T5-11B & 51.8 & \underline{60.3} & 65.0  \\
              \midrule
              \parbox[t]{1.5mm}{\multirow{8}{*}{\rotatebox[origin=c]{90}{FD}}}
               & T5-Large & 59.6 & 61.8 & 64.8\\
               & T5-11B & 70.8 & 62.3 & 65.6  \\
               & GPT3-175B (finetuned) & 75.1 & 64.8 & \textbf{69.8}\\
               & \quad \rotatebox[origin=c]{180}{$\Lsh$} 5-shot & \fiveshotres 57.2 & \fiveshotres 55.1 & \fiveshotres 54.8  \\
               & \quad \rotatebox[origin=c]{180}{$\Lsh$} Zero-shot & \zshotres 51.6 &  \zshotres 56.2 &  \zshotres 55.6  \\
               & GPT 3.5 (5-shot) & 63.8 & 55.6 & 55.2 \\
               & \quad \rotatebox[origin=c]{180}{$\Lsh$} Zero-shot+CoT &  \zshotres 50.4 &  \zshotres 52.8 &  \zshotres  55.4 \\
               & GPT-4 (5-shot) & \textbf{84.5} & \textbf{73.3} & 68.2 \\
               & \quad \rotatebox[origin=c]{180}{$\Lsh$} Zero-shot+CoT &  \zshotres 81.9 &  \zshotres 66.2 &  \zshotres 64.3 \\
               
               \midrule
                &  Human Estimate From Pixels (FP) & 94.0 & 83.7 & 64.6 \\
              \bottomrule
            \end{tabular}
        } %
        \end{minipage}
        \begin{minipage}{0.4\textwidth}
        \centering
        \includegraphics[width=.9\linewidth]{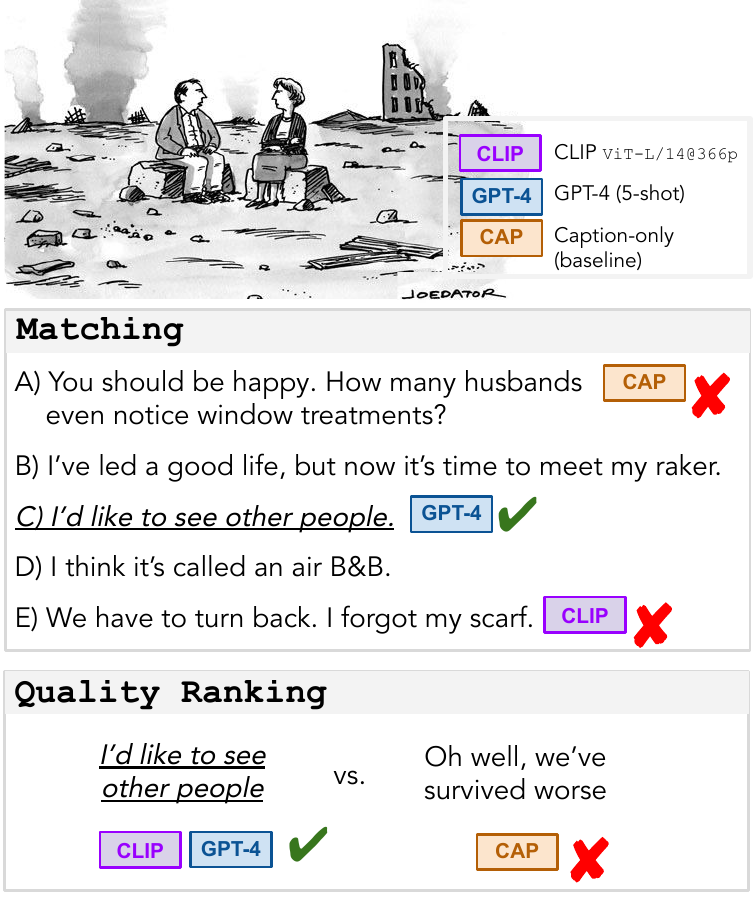}
        \end{minipage}

\caption{Prediction results for the matching and quality ranking tasks: 
averages over 5 cross-validation splits. 
\underline{Underlined} results are the best model in the \emph{From Pixels} (FP) setting, where at test time, models only have access to the cartoon images. \textbf{Bold} results are best in the \emph{From Description} (FD) setting, where at test time, models have access to human-authored descriptions of the cartoons. 
\autoref{sec:results_visualization} presents these results visually. 
Right: sample predictions by CLIP (finetuned), GPT-4 (5-shot), and the caption-only baseline over a matching/ranking instance. Cartoon by Joe Dator.
} %
\label{tab:main_results_no_zeroshot}
\end{table*}

\begin{figure}
    \centering
    \includegraphics[width=.95\linewidth]{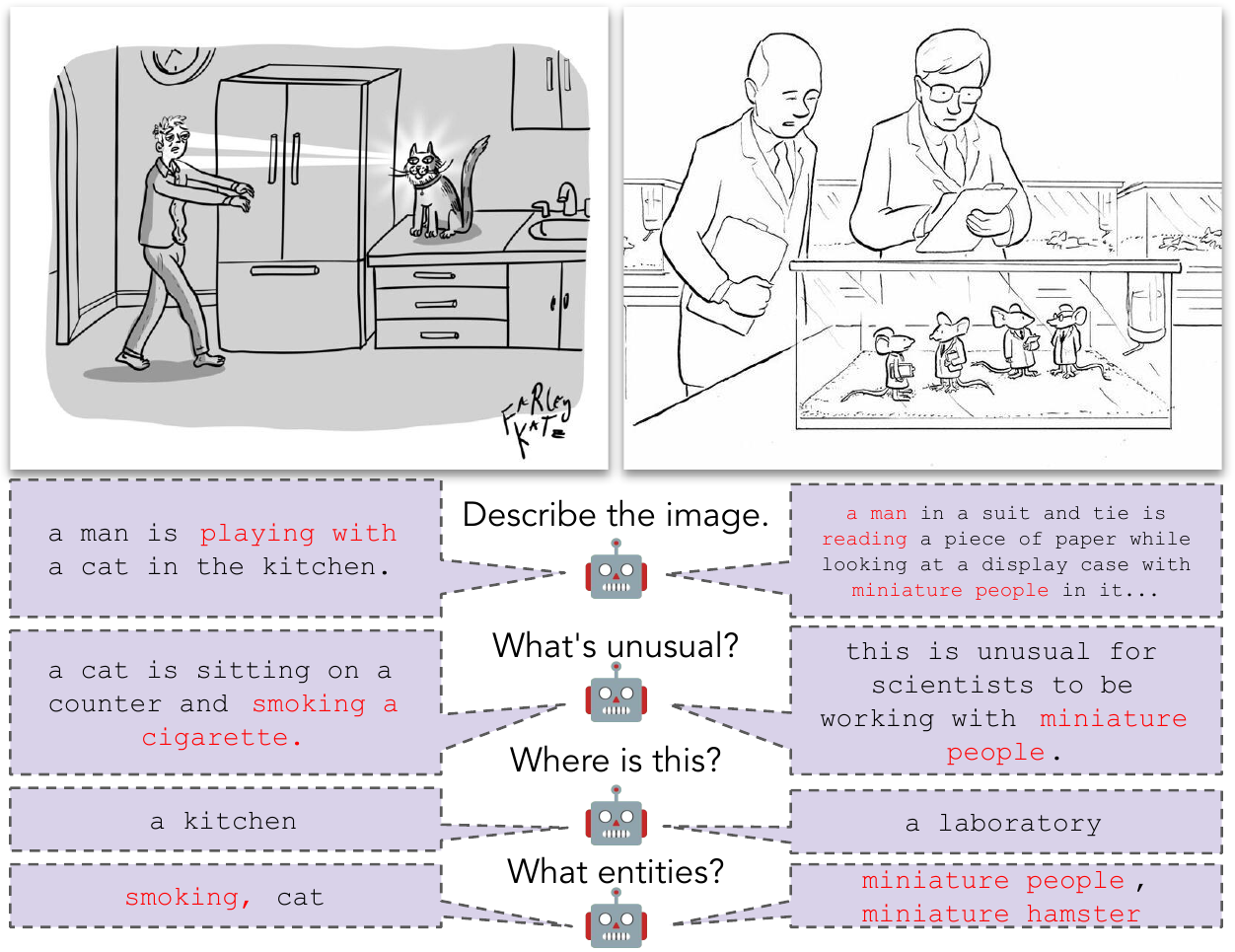}
    \caption{Example predictions by fine-tuned OFA-Huge from images. The model recognizes many objects/actions/locations, but makes some mistakes (indicated in \textcolor{error_color}{red}): for the left image, for example, it 
    falsely indicates that the cat is smoking, and, on the right, that the mice are small people or hamsters (hamsters have stubby tails). %
    Cartoons by Farley Katz and Paul Noth.}
    \label{fig:OFA_predictions}
\end{figure}

\paragraph{OFA $\rightarrow$ LM.} %
We use OFA Huge (930M parameters) \cite{wang2022unifying}, a seq2seq model that supports image/text inputs/outputs; it is pretrained on a variety of vision+language tasks. %
We fine-tune on the New Yorker corpus by training it to map from (cartoon, prompt) $\rightarrow$ descriptions for the %
four types of annotations described in \autoref{sec:sec_with_manual_annotation}; see \autoref{fig:OFA_predictions} for example predictions. %
We organize the OFA-predicted outputs in the same format as the human-authored descriptions in our From Description (FD) models detailed below (except the inputs are the outputs of OFA), and pass the result to a language model:\footnote{We found that fine-tuning OFA directly %
was less effective.} this composition %
can be considered a Socratic Model \cite{zeng2022socratic}.

\subsection*{From Description (FD) Models}

We formulate multiple-choice tasks as text-to-text by concatenating the human-authored cartoon descriptions with the choices %
as input:
the target is simply the letter corresponding to the answer, e.g., \texttt{E}. For \explanation, we autoregressively generate the explanations conditioned on the descriptions/captions.

\paragraph{T5.} We fine-tune T5-Large and T5-11B \cite{raffel2019exploring}; these encoder-decoder transformer models have 770M and 11.3B parameters respectively. For \explanation, we sample with temperature 1.0 and nucleus sampling with p=.95 \cite{holtzman2019curious}.

\paragraph{GPT-3, GPT-3.5, GPT-4.} We use these three OpenAI models as both zero-shot and few-shot models. We provide the models with a description of the task, and, for the few-shot case, 5 random labelled in-context examples. Specifically, for GPT-3 we use \texttt{text-davinci-002} (175B) \cite{brown2020language}, and for GPT-3.5/GPT-4, we use the May 12, 2023 versions \cite{openai2023gpt4}.
For GPT-3, we also consider a fine-tuned version (which is unavailable for GPT3.5/GPT-4).\footnote{\url{https://beta.openai.com/docs/guides/fine-tuning}; for \explanation, we use the default settings; for multiple choice, we set \texttt{prompt\_loss\_weight} to zero. The validation set is not used by the API for early stopping, so we concatenate it with the training set and perform no validation.} For zero-shot GPT-3.5/GPT-4, early experiments revealed that prompting models to ``think'' step-by-step with chain-of-thought (CoT) was helpful \cite{wei2022chain,kojima2022large}.
See \autoref{sec:sec_with_prompting_details} for GPT-3 details, and \autoref{sec:sec_with_new_openai_model_details} for GPT-3.5/GPT-4 details.

\begin{figure*}
\centering
\includegraphics[width=.98\linewidth]{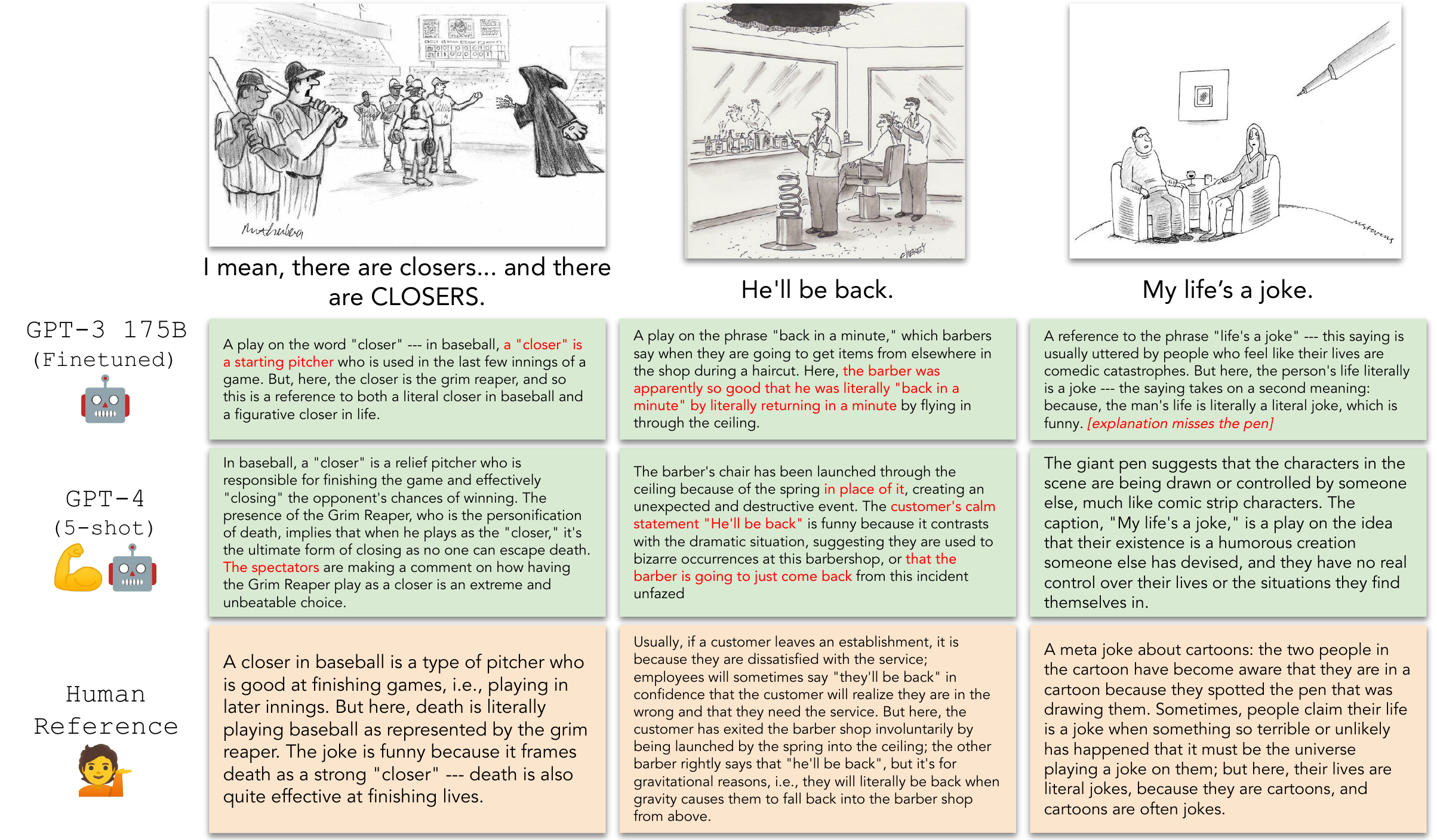}
\caption{A 
random sample of caption explanations generated by a fine-tuned version of GPT-3, 
GPT-4 with 5 shots, and  
human-written references. Errors are highlighted \textcolor{error_color}{in red.} Machine-authored generations are often on the right track, but frequently contain mistakes, e.g., by referring to a closing pitcher as a starter (GPT-3, left image)
or suggesting that a barber, rather than a customer, was launched (GPT-4, middle image).
Cartoons by Mort Gerberg,
Tom Cheney,
and Mick Stevens.
}
\label{fig:predicted_examples}
\end{figure*}

\subsection*{Baselines}

\paragraph{Caption Only.} In addition to a \textbf{Random}-guess baseline, we fine-tune T5-11B given just the caption, i.e., without knowledge of the cartoon \cite{trichelair2018reasonable,poliak2018hypothesis}. %

\paragraph{Human performance estimates.} %
Three people (two authors and one person familiar with the project) each attempted 100 randomly sampled instances from both the \matching and \ranking tasks.\footnote{%
\Matching instances were sampled such that there were no repeated options, i.e., annotators couldn't use process of elimination across instances. 595 total responses were collected.} It is important to note that \emph{%
human performance is not an upper bound for model performance on \matching and \ranking} because labels are not generated by 
a single human and tastes can vary; it can (and does, see \autoref{sec:results}) happen that a machine might be able to reconstruct New Yorker editor preferences more reliably than an untrained human. %
Annotators were given access to the images, but not the descriptions (akin to the FP setting).%

\subsection*{Hardware+software details.}
T5, CLIP, and OFA were trained using 8 A100 GPUs in pytorch \cite{NEURIPS2019_9015}. We use the Transformers \cite{wolf2019huggingface} implementation of T5: T5-11B was trained with deepspeed \cite{rasley2020deepspeed}; T5-Large and CLIP were trained with Accelerate.\footnote{\url{https://huggingface.co/docs/accelerate}}

\subsection{\Matching and \ranking results}
\label{sec:results}
\autoref{tab:main_results_no_zeroshot} contains the results. %
Among the \emph{from description} models, GPT-4 (5-shot) generally performs best, e.g., achieving 84.5\% accuracy on matching. %
It (and fine-tuned GPT-3) also perform
\emph{better} at predicting New Yorker editor selections 
than our three humans (column NYAcc: GPT-3 69.8 vs.~Human estimate, 64.6), but
underperform at 
predicting crowd selections (CrowdAcc column: GPT-4 73.3 vs.~83.7).\footnote{
Also, crowd selectors greatly outnumber New Yorker editors, so crowd rankings may be a more dependable target, statistically speaking.}
We also see that our \emph{from pixels} models leave significant headroom compared to the human performance estimates.

Other observations include: 1) both \emph{from pixels} and \emph{from description} models 
mostly
outperform the Caption Only baseline (even for smaller model sizes), suggesting that the models are truly using feature interactions between cartoons/captions to improve their predictive accuracy; 2)  fine-tuning CLIP tends to do best for \matching in the  \emph{from pixels} setting, but OFA+T5-11B is competitive for \ranking (and supports generation, see \autoref{sec:sec_with_pairwise_comparisons}); and 3) the performance difference between T5 vs.~OFA$\rightarrow$T5 
exemplifies
the effect of suboptimal visual recognition when shifting from the \emph{from pixels} setting to the \emph{from description} setting. Finally, while performance drops are incurred universally for zero-shot models, %
pointing towards the utility of the new annotated corpus we are releasing (\autoref{sec:sec_with_manual_annotation}), GPT-4's zero-shot chain-of-thought incurs a smaller performance drop compared to other zero-shot models; see \autoref{sec:sec_with_new_openai_model_details} for a sample chain-of-thought.

\subsection{Human evaluation of \explanation.}

\label{sec:sec_with_pairwise_comparisons}

We %
gather judgments from 3 crowd-workers per test instance by asking them which of a pair of explanations they prefer, and take a majority vote to determine a winner.  Results and annotator agreement are in~\autoref{tab:pairwise_human_evaluations}, and samples of GPT-3, GPT-4, and human joke explanations are in \autoref{fig:predicted_examples}.  Our evaluations address seven questions:
\paragraph{Q1: Do models utilize the image context of the caption to generate better explanations?}
\emph{Test: T5-11B vs.~Caption-only T5-11B.}
Answer: \textbf{Yes.}
Compared to the same model trained with no access to image information, the model with image information wins in 84.7\% of cases.

\paragraph{Q2: 
Is computer vision a bottleneck
for top-quality explanation generation?}
\emph{Test: T5-11B  (in the FD setting) vs.~OFA $\rightarrow$ T5-11B.}
Answer: \textbf{Yes.}
Compared to the same model trained with access to human written descriptions available at test time (i.e., the \emph{from description} setting), the model trained with access only to OFA-predictions loses in 74.6\% of cases.

\paragraph{Q3: Do bigger T5 models generate better explanations?}
\emph{Test: T5-11B vs.~T5-Large.}
Answer: \textbf{Yes.}
T5-11B with access to the same information at test time as T5-Large (770M) is preferred in 68.5\% of cases.

\begin{table}
     \resizebox{.99\linewidth}{!}{
     \centering
     \begin{tabular}{cll|ccc}
     \toprule
     & A & B & \% A wins & \# ratings & G-$\gamma$ \\
     \midrule
     Q1 & T5-11B & Caption only & 84.7\% & 393 & 64.4 \\
     Q2 & T5-11B & OFA $\rightarrow$ T5-11B & 74.6\% & 393 & 41.6 \\
     Q3 & T5-11B & T5-Large & 68.5\% & 390 & 45.9 \\ %
     Q4 & FT-GPT-3 & In context GPT-3 & 50.0\% & 396 & 23.2\\
     Q5 & 5-shot GPT-4 & Zero-shot GPT-4 & 64.3\% & 396 & 19.7 \\
     Q6 & 5-shot GPT-4 & 5-shot GPT-3 & 93.0\% & 384 & 86.4 \\
     Q7 & Human & 5-shot GPT-4 & 67.7\% & 390 & 20.9 \\
     \bottomrule
     \end{tabular}
     }
     \caption{Pairwise human evaluations for \explanation, with per-instance agreement according to Gwet's (\citeyear{gwet2001handbook}) $\gamma$. Q1-Q7 notations refer to the corresponding paragraphs in \autoref{sec:sec_with_pairwise_comparisons}.
     }
 \label{tab:pairwise_human_evaluations}
 \end{table}

\paragraph{Q4: Does fine-tuning an LLM model help vs. in-context learning for explanation generation?}
\emph{Test: FT-GPT3 vs.~In context (=5-shot) GPT3.}
Answer: \textbf{Not really.}
In contrast to the multiple choice tasks, %
we find that in-context explanation
generations are comparable to fine-tuned ones according to pairwise human evaluations, even though the perplexity of the in-context model, reported in \autoref{sec:sec_on_auto_eval},  is much higher (107 vs.~21.8).\footnote{A disparity not mirrored in the word-overlap metrics BLEU-4 and Rouge-L, also reported in \autoref{sec:sec_on_auto_eval}.} We expect that the fine-tuned model more closely mirrors the style of the corpus, but that the in-context explanations also contain similar content, e.g., relevant entities.

\paragraph{Q5: Do supervised explanations help, even with GPT-4?}
\emph{Test: 5-shot GPT-4 vs.~Zero-shot GPT-4.}
Answer: \textbf{Yes.} The zero-shot version of GPT-4 is missing access not only to the supervision of paired (caption, explanation) data, but also, explanations in the detailed style of our released corpus. Perhaps as a result, 5-shot GPT-4 (which also achieves significantly higher BLEU-4/Rouge-L)
is preferred in 64\% of cases.

\paragraph{Q6: Does GPT-4 outperform GPT-3?}
\emph{Test: 5-shot GPT-4 vs.~5-shot GPT-3.}
Answer: \textbf{Yes, definitely.} In our most definitive result, with equal amounts of supervision, GPT-4's explanations are preferred nearly universally --- specifically, in \emph{93\%} of cases. Interestingly,  GPT-3 performs slightly better on automatic evaluation metrics for explanation like BLEU-4 and Rouge-L (see \autoref{sec:sec_on_auto_eval}), which suggest that the earlier family of may fit the surface features of the generation task more effectively, e.g., 5-shot GPT-3 achieves 5.07 BLEU-4 compared to 4.99 for 5-shot GPT-4. This suggests that mirroring the surface form of our explanation corpus is not sufficient to generate the highest quality explanations.

\paragraph{Q7: Does our best model, GPT-4, explain jokes as well as humans?}
\emph{Test:  Human vs. Few-shot GPT-4.}
Answer: \textbf{No.}
Human-written explanations are preferred by annotators in 68\% of pairwise cases.\footnote{For a similar, earlier set of experiments with FT-GPT-3 vs. human, human was preferred in 87.8\% of pairwise cases.}
We qualitatively examine the 39/130 cases where the human reference receives 3/3 annotator votes. In these cases, the machine-generated explanations usually incorrectly interpret the image, e.g., in one case, a caption jokes about two cavepeople in a hole looking at a caveman in a cave with the caption ``Personally, I'm not a big fan of modern architecture."; GPT-4 incorrectly interprets the hole as ``modern architecture'' instead of the cave. %
We also examine the 8/130 cases where the GPT-4 produced caption was unanimously preferred: a close reading of these cases is provided in \autoref{sec:sec_with_machine_preferred_explanations}. 
In 3 of these 8 cases, the human explanations, while on the right track, had slight inaccuracies, and in the remaining 5 cases, the human and machine explanations both express the same idea, but with different styles (GPT-4's sometimes arguably being more formal, detailed, or fluent).

\subsection{Error Analysis for \Matching}

We conduct an error analysis of a performant \emph{from pixels} model (\texttt{CLIP ViT-L/14@336px} finetuned), and a performant \emph{from description} model (GPT3-175B finetuned). We concatenate the test set predictions over the 5 cross validation splits, and ask: %

\paragraph{Q8: Are some contests more difficult than others?}
Answer: \textbf{Yes.}
\emph{Details:} We conduct a $\chi^2$ test by forming a contest-by-correctness (704-by-2) contingency table, aggregating over the 3-6 \matching instances for each contest, and find that errors are clustered according to contest ($p<.05$ for both CLIP and GPT-3).\footnote{Similar $\chi^2$ tests find no evidence of correlation between correctness and (a) cross-validation split (5-by-2 table; p=.84/.14 for GPT3/CLIP); or (b) which captions are randomly assigned as negative choices (2646-by-2 table, p=.92/.79 for GPT3/CLIP).} There's a moderate Spearman correlation between the per-contest accuracy between the models ($\rho=.28, p\ll.001$), but (as a null hypothesis) only a slight correlation between contest date and difficulty for either (later contests easier, GPT3/CLIP $\rho=.07/.08, p=.08/.05$). When the models' predictions agree, they are correct 87\% of the time. When GPT-3 is wrong, CLIP is right only 38\% of the time; under the null hypothesis that their errors are uncorrelated, CLIP's accuracy would be 62\% ($p \ll .001$ errors are uncorrelated, permutation test).
However, when we attempt to identify consistent factors that predict contest difficulty using various visual/linguistic predictors, we find  hard vs.~easy difficult to predict \emph{a priori}; our best classifiers perform only slightly above random. We will distribute the hard vs. easy contest lists as a resource for future work.

\section{Related Work}

\paragraph{Humor.} \newcite{raskin1979semantic} and \newcite{attardo2008primer} highlight three
``great families'' of theories of the roots of humor: 1) \emph{hostility,} 
claims of superiority over someone or something 
 \cite{gruner1978understanding,billig2005laughter}; 2)
\emph{release} %
of a constraint
\cite{freud1905jokes,fry1963sweet,mindess1971laughter} and
3) \emph{incongruity,} %
\cite[sometimes ``incongruity-resolution'';][]{mulder2002humour} %
the
introduction (and subsequent resolution) of generally
incompatible contexts %
\cite{schopenhauer1891world,shultz1976cognitive}. \newcite{shahaf2015inside} note that most New Yorker caption contest cartoons involve incongruous situations.%

\paragraph{NLP + The Caption Contest.}
\newcite{king2013random},
\newcite{shahaf2015inside}, and
\newcite{radev-etal-2016-humor} analyze 5, 16, and 50 New Yorker Caption Contests, respectively. 
Best-performing features for identifying the funniest among a set of caption choices include: perplexity, match to image setting and uncanniness description, readability, proper nouns  \cite{shahaf2015inside},  overlap with WordNet's \citep{fellbaum:98a} ``person'' and ``relative'' synsets,  lexical centrality among submissions  \cite[inspired by \citet{mihalcea+pulman:2009}]{radev-etal-2016-humor}, and sentiment (both papers).
Our ``location'' and ``uncanny description'' annotations are direct analogs of the ``context'' and ``anomaly'' tags of \newcite{shahaf2015inside}, and our data incorporates that generously released
by the previous researchers. Our extensions are (a) the addition of two novel tasks; 
(b) using new data/resources/models to curate ranking pairs (see \S\ref{sec:sec_with_original_corpus_details}); and (c) evaluating two distinct audience preferences: New Yorker editors vs.~``the crowd''. \autoref{sec:sec_with_beyond_peer_reviewed_work} highlights efforts beyond the scope of peer reviewed AI venues, e.g., blog posts. %

\paragraph{Measuring preferences over captions.}
While humor is ultimately subjective, work on the contest has studied modeling \emph{average} preferences of raters.
\newcite{NIPS2017_c02f9de3} design quality ranking algorithms for the caption contest, framed as identifying the best ``arm'' in a multi-armed bandit setting; their crowdsourcing system NEXT \cite{jamieson2015next} is  \href{https://www.newyorker.com/cartoons/vote}{used by The New Yorker}. It does not directly use the content of the cartoons/contests. The result is \newcite{newyorkernextmldataset}'s %
continuously updated corpus, 
from which we draw some of our data.

\paragraph{Multimodal and computational humor.} %
\newcite{chandrasekaran2016we} explore humor recognition in images, and \newcite{castro-etal-2019-towards,hasan2019ur,patro2021multimodal,hasan2021humor} explore laughter prediction in TED-talks/sitcoms. \citet{tsakona2009language,fallianda2018analyzing} study political cartoons.
\citet{chakrabarty2022flute} recently proposed a version of NLI for figurative language, which can be humorous. %
Some work has tried to detect whether a sentence is humorous or not \cite{blinov2019large,annamoradnejad2020colbert}.
More difficult to evaluate \cite{valitutti2011many} are setups where the goal is to automatically generate humorous content in various contexts
\cite{binsted1994implemented,stock2003getting,mihalcea2005making,mihalcea2006technologies,wang2015can,chandrasekaran2017punny,yoshida2018neural,sundaram2018generation,shimomoto2019news2meme}; a survey is provided by \newcite{amin2020survey}.

\paragraph{Explaining humor.} In the taxonomy of \newcite{tan2021diversity}, joke explanations are most related to proximal mechanisms:
``This type of explanation attempts to provide the mechanism behind the predicted label, i.e., how to infer
the label from the text'', or efficient cause a la Aristotle \cite{lombrozo:2006a}. \citet{chowdhery2022palm} undertake a qualitative exploration of (non-visual) joke explanations.

\section{Conclusion}

\label{sec:sec_with_human_AI_collaboration_discussion}
We demonstrate that today's vision and language models still cannot recognize caption relevance, evaluate (at least in the sense of reproducing crowd-sourced rankings), or explain The New Yorker Caption Contest as effectively as humans can.
However, the partial capacity of today's AI is still substantial, and may be sufficient for models to serve as creative collaborators, e.g., as brainstorming assistants for humorists/cartoonists. Specifically: 1) our \matching/\ranking models could help entrants receive quantitative feedback on the relevance/predicted quality of their submissions, and 2) the annotated corpus+explanations we introduce could be repurposed for generation
(we explore generation of novel cartoons/captions in~\autoref{sec:sec_with_human_AI_collaboration}). Finally, a promising avenue for future work focused on generating humorous captions (c.f.\ our focus of humor ``understanding" benchmarks) would be to operationalize the feedback provided by our matching/ranking models in an reinforcement learning from human feedback (RLHF) loop.

\paragraph{A last remark.}  We cannot claim to know whether the human-machine `humor understanding gap' will be closed sooner or later.\footnote{Or never.  \href{https://www.newyorker.com/cartoons/bob-mankoff/the-story-of-how-about-never}{Is never good for you?}} But %
we encourage other researchers to have as much fun with the topic as we did! %

\section{Limitations}

The New Yorker Cartoon Caption Contest represents a narrow slice of humor, deriving from a particular language, region, history, culture, style, and set of conventions. Hence, the results of this study do not represent or cover {all} types of humor. 

Our framing of the \ranking task could be interpreted as seemingly prescriptive (i.e., that joke A is ``objectively'' better than joke B), but New Yorker editorial selections should not be taken as ground truth for funniness; disagreement about what is funny is expected and valid. Our tasks operationalize the prediction of only \emph{average} preferences (rather than individual ones), and these preferences may include a partiality or bias towards items that conform to the characteristics of prior contest winners or published New Yorker cartoons.

Finally, the explanations in our annotated corpus were largely written by a single author of this paper.  While a larger pool of the crowdworkers judged these explanations to be of higher quality in comparison to machine generations, future work would be well-suited to compare the person-to-person variance in explaining why particular jokes are funny.

\section{Acknowledgements}
We thank the cartoonists and contest entrants for their wonderful efforts!
We additionally thank our crowd annotators for their diligent work, Lisa Watkins for contributing to the human performance estimates, 
and the anonymous reviewers for their constructive comments.
This work was funded in part by DARPA MCS through NIWC Pacific (N66001-19-2-4031), the Allen Institute for AI,
and a Google Focused Research Award.
Jack Hessel conducted initial work while at Cornell University. Ana Marasovi\'{c} conducted this work while at The Allen Institute for AI. Rowan Zellers conducted this work while at University of Washington.

\bibliography{refs}
\clearpage
\appendix

\section{Crowdworking Details}

\label{sec:sec_with_crowdworking_details}

\begin{figure*}
    \centering
    \includegraphics[width=.49\linewidth]{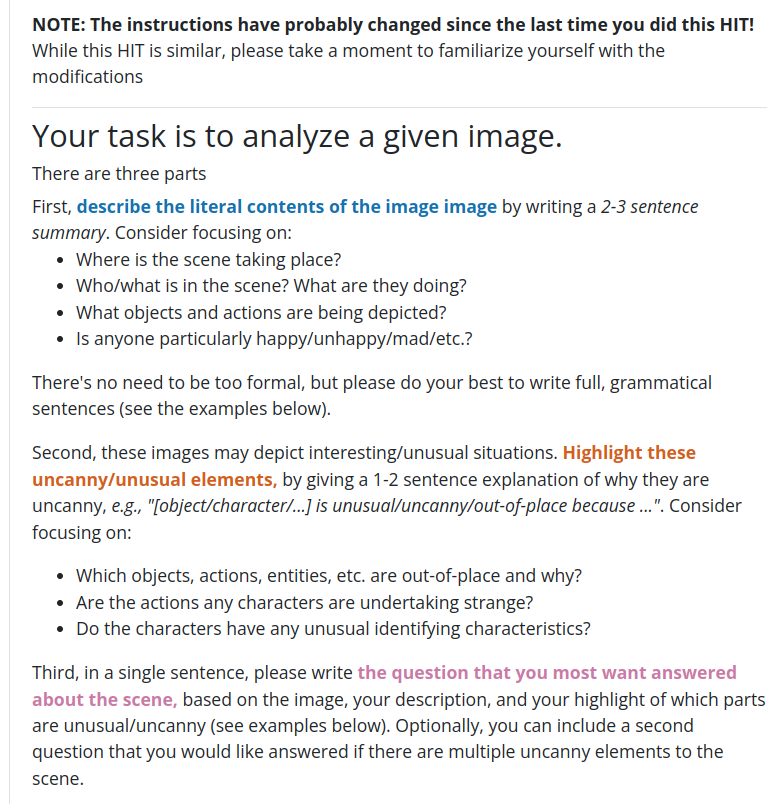}
    \includegraphics[width=.49\linewidth]{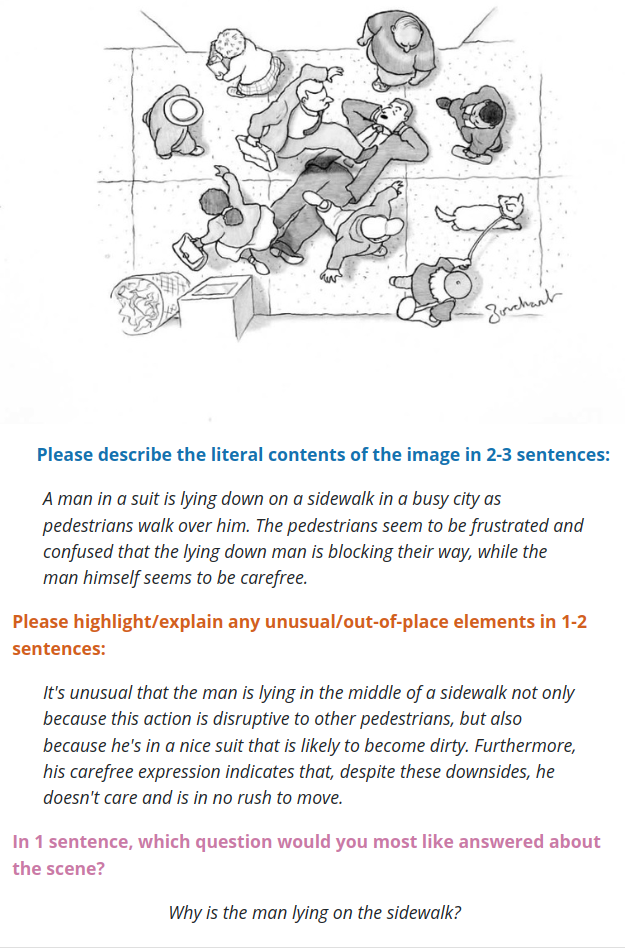}
    \includegraphics[width=.35\linewidth]{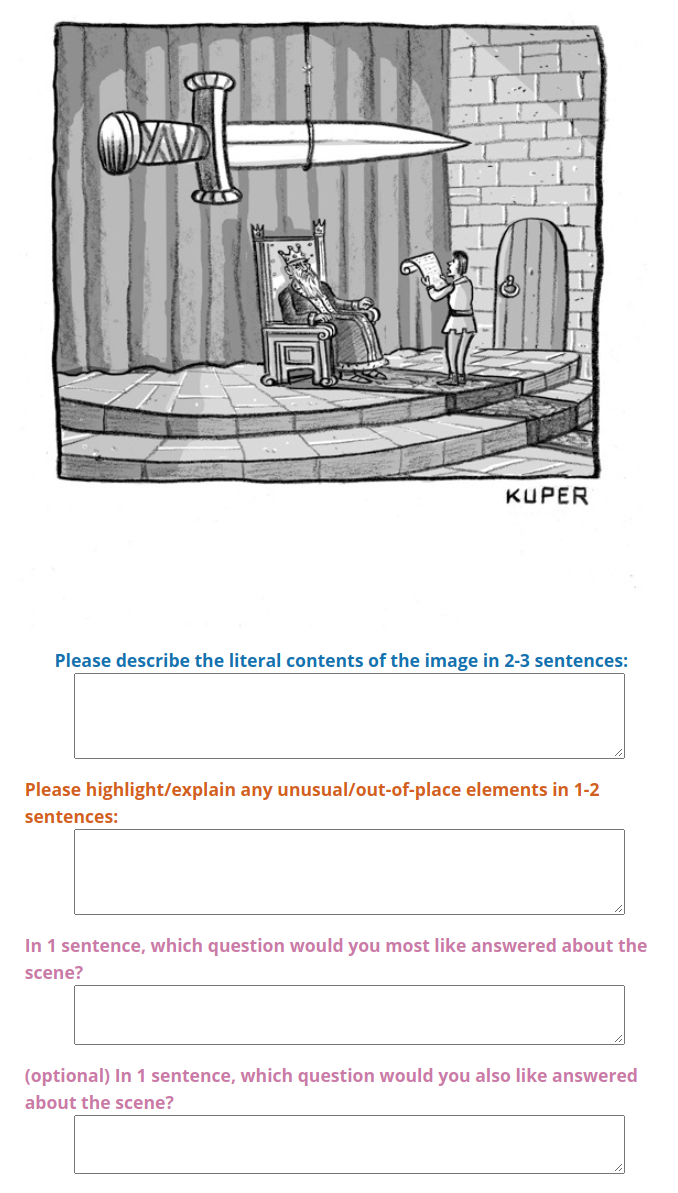}
    \caption{Instructions, example, and interface for the Cartoon Description HIT. We gather, but do not use, the final ``Which question?" annotation in our experiments.}
    \label{fig:description_interface}
\end{figure*}

\begin{figure*}
    \centering
    \includegraphics[width=.49\linewidth]{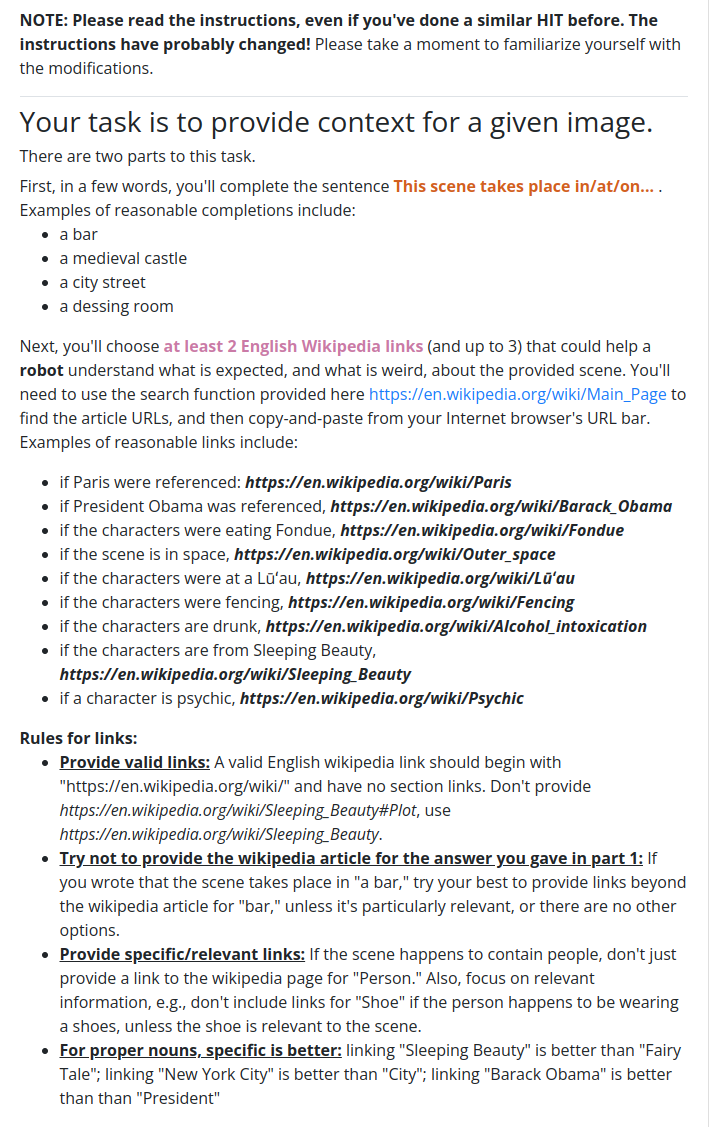}
    \includegraphics[width=.49\linewidth]{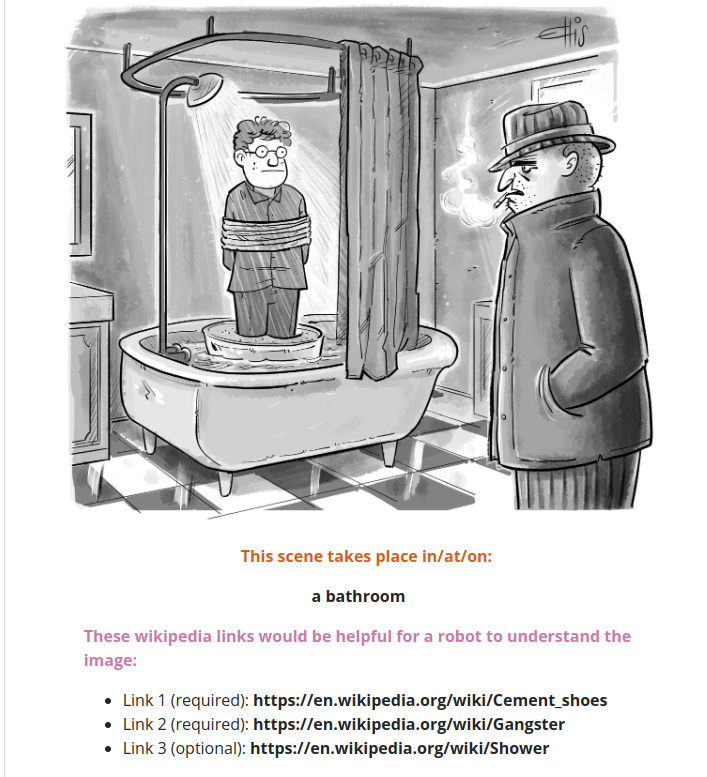}
    \caption{Instructions and example for the Wikipedia links HIT.}
    \label{fig:link_interface}
\end{figure*}

\begin{figure*}
    \centering
    \includegraphics[width=.49\linewidth]{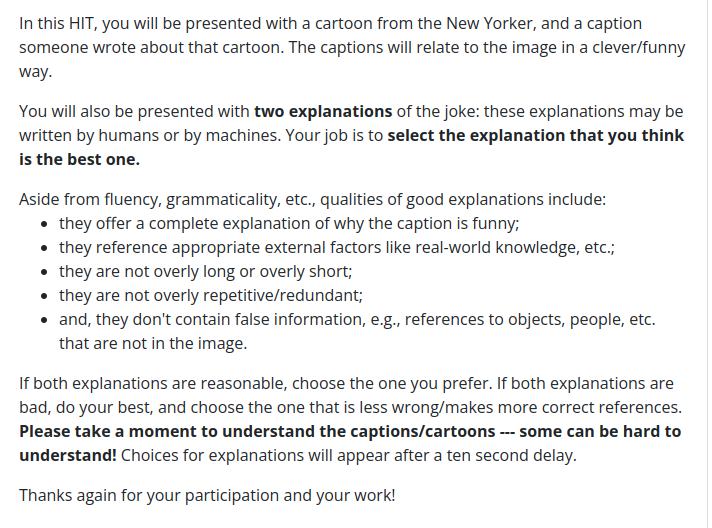}
    \includegraphics[width=.49\linewidth]{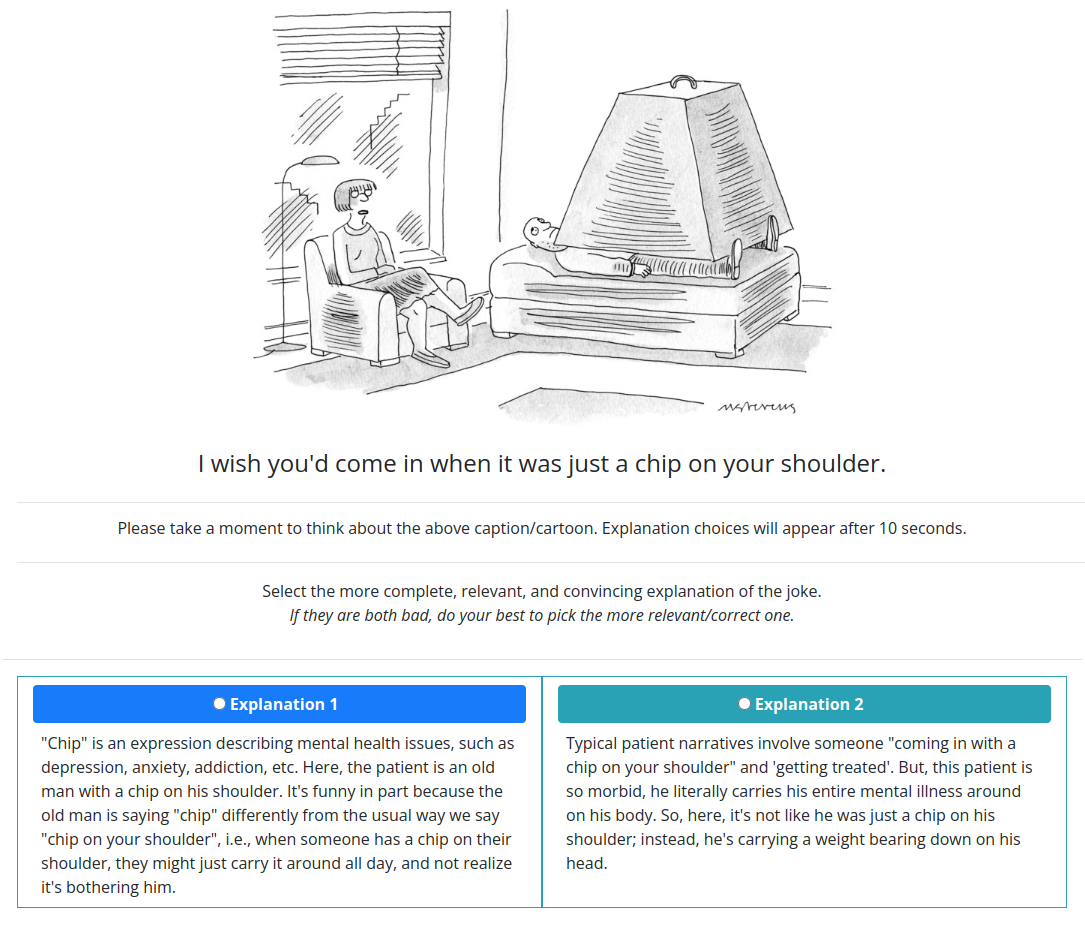}
    \caption{Instructions and interface for the pairwise explanation judgment HIT.}
    \label{fig:pairwise_interface}
\end{figure*}

We use three Mechanical Turk interfaces to gather data. These are:

\begin{enumerate}[leftmargin=*,topsep=0pt,itemsep=-1ex,partopsep=1ex,parsep=1ex]
\item \emph{Cartoon description} (\autoref{fig:description_interface}). We ran this HIT 3 times per cartoon.
\item \emph{Cartoon wikipedia links} (\autoref{fig:link_interface}). We ran this HIT 2 times per cartoon.
\item \emph{Pairwise explanations} (\autoref{fig:pairwise_interface}). We ran this HIT 2.7K times to facilitate the comparisons in \S\ref{sec:sec_with_pairwise_comparisons}
\end{enumerate}

\paragraph{Qualification+training rounds.} To ensure our set of crowdworkers were properly trained for the annotations, we ran two types of qualification rounds: one for the description/link HITs, and one for the pairwise explanation HITs.

For the description/link HITs, our qualification round was based off an earlier and more involved HIT that involved a joint setup where, for 3 cartoons, users described cartoons, highlighted image regions, explained jokes, etc. We allowed users from $\{$AU, CA, NZ, GB, US$\}$ with 10K prior approved HITs and a minimum acceptance rate of 97\% on their previous HITs to participate. Some of the cartoons and captions contain mature themes; we provided the recommended disclaimer for this and other HITs: ``WARNING: This HIT may contain adult content. Worker discretion is advised." We manually graded the responses of 30 annotators in a qualification round, and qualified 21. Through a mix of the older, more involved HITs and the streamlined HIT in \autoref{fig:description_interface}, which is a pared-down version of the original HIT without captions, we gathered  descriptions of the cartoons. We also gathered the locations/Wikipedia entity links from the qualified annotators. These annotations were gathered in mid-to-late 2021.

About 9 months later, we conducted a second set of Mechanical Turk studies for pairwise judgment evaluations for \explanation. A second qualification round was run, in which we asked annotators to rate the quality of several joke explanations which we manually selected to be good/bad across various desirable axes. We qualified 29 out of 51 annotators who attempted the HIT via manual inspection of their responses. This set of annotators were given access to the final pairwise-judgment HITs.

Crowdworking studies of standard computer vision corpora (involving no personal disclosures) are not required by our IRB to be reviewed by them.
While the authors of this work are not lawyers and this is not legal advice,
this opinion is based on United States federal
regulation 45 CFR 46, under which this study qualifies and as exempt. We hashed crowdworker IDs in the public release so annotations cannot be back-traced to individual workers.

\section{Additional Experimental Details}

\label{sec:sec_with_additional_experimental_details}

\subsection{From Description details} For each cartoon, we have multiple annotations of each type, as detailed in \autoref{sec:sec_with_manual_annotation}. During training, we utilize all location/description/uncanny description/sets of links, but at test time, we randomly sample a single set of these four annotation types such that inference requires only a single forward pass. For fair comparison, the randomly sampled description available at test time is held constant between all methods.

More detail about how we managed multiple annotations: because we have 2 locations $\times$ 3 descriptions $\times$ 3 uncanny descriptions $\times$ 2 entity links, there are potentially 36 possible combinations we could use to form a \emph{from description} instance for each cartoon. However: tuples are constructed at the annotator level to account for potential dependencies between annotation types: because descriptions/uncanny descriptions were were collected in the same HIT, the uncanny description may reference entities from the description because they were authored at the same time by the same annotator in sequence. Similarly, the (locations, links) were collected in the same HIT. So, we instead consider all six possible tuples holding author constant between HITs, i.e., 3 (description, uncanny description) $\times$ 2 (location, link). For test time, select a single random valid tuple of annotations for evaluation, which is fixed for all comparisons.

\subsection{CLIP} 
For fine-tuning results, we do linear warmup for 200 steps and conduct a small learning rate search on the validation set for each cross-validation split independently between $\{5e\mhyphen5,1e\mhyphen5,5e\mhyphen6\}$, keeping batch size fixed at 32. To keep the entire cartoon in the 336px square input, we resize and pad. At training time, we perform data augmentations on the image, including: random horizontal flipping, random color jittering, and random grayscaling.

\subsection{OFA} \label{sec:OFA_training_details}
We use validation-set early stopping on cross-entropy loss, and fine-tune OFA separately for each cross-validation split. After fine-tuning, we select the top-1 prediction according to beam search (n=5).
We finetune OFA Huge with a learning rate of $5e\mhyphen5$, which was determined via a small grid search over the first cross-validation split. We use label-adjusted smoothed cross entropy loss as implemented by the OFA authors\footnote{\url{https://github.com/OFA-Sys/OFA}} with smoothing of 0.1. We train for a maximum of 7 epochs with a warmup ratio of 6\%. For each image, we query for the four different types of annotations shown in~\autoref{fig:annotations}. To facilitate this, in addition to providing OFA with the image, we also provide it with a per-annotation-type prompt:

\begin{enumerate}[leftmargin=*,topsep=0pt,itemsep=-1ex,partopsep=1ex,parsep=1ex]
\item for locations: ``Where does this take place?"
\item for descriptions: ``Describe this image."
\item for uncanny: ``What's unusual about this image?"
\item for entities: ``What entities are there?"
\end{enumerate}

In early experiments, instead of composing with a language model, we did attempt to fine-tune OFA directly for the \explanation task. However, we found that the resulting perplexity (roughly 300) was significantly higher than for other fine-tuned models, with the errors difficult to diagnose.

\subsection{T5-Large/T5-11B.} For T5-Large, we conduct a small, per-cross-validation split learning rate search between $\{1e\mhyphen4, 1e\mhyphen5, 5e\mhyphen5\}$ and keep batch size fixed at 64.  For T5-11B we use a fixed learning rate of $1e\mhyphen5$ and a batch size of 64.

\subsection{GPT-3 Zero Shot/In Context}
\label{sec:sec_with_gpt3_details}
 We use GPT-3's \texttt{davinci-text-002} model for our main zero shot and in-context learning experiments. Examples of zero-shot prompts for all tasks are given in  \autoref{fig:zero_shot_prompt_examples}. The in-context prompts are similar, except they contain 5 random samples from the training set. A full, randomly selected in-context prompt for the explanation generation task is given in \autoref{tab:full_prompt_example}.

\begin{figure*}
\centering
\resizebox{.95\linewidth}{!}{

\begin{tabular}{p{2\linewidth}}

\toprule
In this task, you will see a description of an uncanny situation. Then, you will see five jokes --- only one of which was written about the described situation. Pick which of the five choices truly corresponds to the described scene.\\
\#\#\#\\\\

This scene takes place in the following location: boardroom. Four birds are in an office. They're perched around a table. Birds don't have offices. The scene includes: Parrot, Speech\_repetition.\\
one of the following funny captions is most relevant to the scene:\\\\

A) Just be glad he's not wearing his kilt today.\\
B) The founding fathers were clear. You must win by two.\\
C) She'll appreciate you're wearing protection.\\
D) We have to stop eating the seed money.\\
E) Can I interest you in opening an offshore account?\\\\

the funny caption that matches the scene is:\\
\midrule

In this task, you will see a description of an uncanny situation. Then, you will see two jokes that were written about the situation. One of the jokes is better than the other one. Pick which of the two jokes is the one rated as funnier by people. \\
\#\#\#\\\\

This scene takes place in the following location: a cave. A caveman is drawing a picture of an elephant on his cave wall. The elephant is standing by as a model. The elephant is friends with a man. The scene includes: Caveman, Mammoth, Cave\_painting. choices:\\
A) Trust me. One day your portrait will be used as the symbol of a political party even more primitive than we are.\\
B) So I've added the pointy trunk. Were there any other unique characteristics the mugger had that you remember?\\\\

the funnier is: \\
\bottomrule 

\end{tabular}
}
\caption{Example GPT-3 zero-shot prompts for \Matching (top) and \Ranking (bottom) tasks. In-context prompts are similar, except 5 random labelled training examples are also provided in the prompt.}
\label{fig:zero_shot_prompt_examples}
\end{figure*}

\begin{figure*}
\centering
\resizebox{.95\linewidth}{!}{
\begin{tabular}{p{2\linewidth}}

\toprule

In this task, you will see a description of an uncanny situation. Then, you will see a joke that was written about the situation. Explain how the joke relates to the situation and why it is funny.\\
\#\#\#\\\\

This scene takes place in the following location: a laboratory. A man in lab attire is sitting in a room with several monkies. Most are in cages, but one of them is in front of a typewriter It's unusual to see a monkey operating a typewriter The scene includes: Infinite\_monkey\_theorem, Mad\_scientist.\\
caption: Have you considered writing this story in the third monkey rather than the first monkey?\\
explanation of the caption:\\
Stories can be told in first person (e.g., ``I did X") or third person (``The character did X"), and editors will sometimes critique this choice or offer suggestions about the writing style. But here, the monkey is writing, so instead of first/third person, the suggestion about perspective is first/third ``monkey".\\
\#\#\#\\
This scene takes place in the following location: city. Devils are herding people in the middle of a street. A manhole is open and there is fire below. Devils in the middle of a city are pretty out of place, and hell being in the sewers is too. The scene includes: Devil, Sewerage.\\
caption: Watch your step, I think this street is paved with Good Intentions.\\
explanation of the caption:\\
A play on the figurative saying ``The Road to Hell is Paved with Good Intentions" --- this saying means that good intentions, left unchecked, can often lead to the worst outcomes. But here, the street is literally a path to hell, and so the man is saying it must be the street from the saying.\\
\#\#\#\\
This scene takes place in the following location: a streetcorner. Two angels driving a police car look on as two other angels loiter and drink on a street corner. The scene is cloudy, which implies that this is happening in heaven Heaven is an idealistic place where one wouldn't expect police to be needed. Furthermore, the folks sitting on the sidewalk and drinking do not accord with a the platonic, perfect view of heaven --- if things are so perfect, they probably wouldn't need to be drinking. The scene includes: Heaven, Public\_intoxication, Angel.\\
caption: Except here it's always good cop, good cop.\\
explanation of the caption:\\
A play on the phrase ``good cop, bad cop" --- two homeless angels are loitering in heaven as two angel cops come by. Instead of the usual ``good cop, bad cop" routine where one officer pretends to be nice and the other is mean as a strategy to get a confession, in heaven apparently all cops are good, so it's ``good cop, good cop".\\
\#\#\#\\
This scene takes place in the following location: an office. Two people are sitting across a desk from one another. They are both tied up in straight jackets. People don't normally wear straight jackets in offices. The scene includes: Straitjacket, Interview.\\
caption: I have some good news—we're letting you go.\\
explanation of the caption:\\
A play on the words ``letting you go" -- normally, if your boss was to ``let you go" it would be bad news, because they would be firing you. In this context, ``letting you go" means removing you from your straight jacket, which would be good news. Because the boss also has a straitjackets, the employee might be getting fired from this company where everyone wears straitjackets, but that would still be good news because this is a terrible workplace.\\
\#\#\#\\
This scene takes place in the following location: a desert. There is a man rowing a boat across a sandy desert who is looking down at another man next to his boat. The other man is crawling on all fours and is looking up and talking to the man in the boat. It is unusual that a man is rowing a boat through sand. It is also unusual to find a man crawling through a desert on all fours. The scene includes: Rowing, Thirst.\\
caption: Long time no sea.\\
explanation of the caption:\\
A play on the term ``sea" --- ``long time no see" is something people say to each-other after they meet up for the first time in an extended period. But here, "sea"  transforms the meaning to mean: ``it's been a long time since you've been in the ocean," which is an accurate comment for someone who has been paddling a boat through the desert for many miles.\\
\#\#\#\\
This scene takes place in the following location: an office. There are three man in the room of an office building. One man is carefully trying to place files inside a filing cabinet. There is a tiger on a leash sitting on top of the filing cabinet. There is a wild tiger inside of an office building. The scene includes: Filing\_cabinet, Tiger.\\
caption: You know, I'm gonna miss this when we go paperless.\\
explanation of the caption:\\
\midrule
\texttt{text-davinci-002} \\
\begin{flushright}  The man is trying to put files into a filing cabinet, but there is a tiger sitting on top of it. The tiger is probably taking up a lot of space and making it difficult to put the files away. The man says he's going to miss this when they go paperless, meaning he's going to miss the struggle of trying to put files away with a tiger in the way, because once they go paperless there will be no more files to put away. \end{flushright} \\
\begin{flushright} This is a joke about how going ``paperless" (i.e., using digital files instead of physical paper) is often touted as being more environmentally friendly. But here, the man is saying he's going to miss having a tiger in the office when they switch to digital files --- presumably because it won't be as exciting. \end{flushright} \\
\begin{flushright} A play on the term ``paperless" --- ``going paperless" is a goal for many companies because it would save money and be more efficient. But here, the company is so far from that goal that they have a tiger on top of their filing cabinet. So instead of ``going paperless," this company is going ``tiger-full." \end{flushright} \\
\midrule
\texttt{text-curie-001} \\
\begin{flushright} A play on the phrase ``going paperless" --- instead of carrying around a bunch of folders, documents, and other paperwork, people might move to a system where everything is electronic. But here, the tiger is a reminder that not everything will be digital, and there will still be times when people need to file things away in a physical filing cabinet. \end{flushright} \\
\begin{flushright} A play on the phrase "going paperless" --- in the past, when people had to file paper documents, it would take up space and be a hassle to manage. But now, with electronic files, the phrase "going paperless" means that all of these documents will be stored electronically, and so they will no longer take up space or be a hassle to manage. However, in this situation, the tiger will still be around, meaning that the transition to a paperless system has not been completely successful. \end{flushright} \\
\begin{flushright} A play on the phrase "going paperless" --- when people go paperless, they means they're going to stop using filing cabinets, which is a physical representation of the transition to a digital world. In this scenario, the joke is that the man is going to miss the tiger, which is a physical representation of the tiger being inside of the filing cabinet. \end{flushright} \\
\midrule
\texttt{text-babbage-001} \\
\begin{flushright}  A play on the phrase ``paperless" --- in a paper-based office, the files would be placed on a paper filing cabinet, which would be taken away when the office moved to a paperless setting. The tiger might symbolize the fear of not being able to find or eliminate the files, as well as the idea of being lost in an office with no way to find your way out. \end{flushright} \\
\begin{flushright} A play on the words ``paper" and `` loneliness." It is interesting to see the tiger in the office, because it is a reminder of the files that the man is trying to place inside the cabinet. In the same way, the tiger is a reminder of the man who is lost and lonely, and the office is a place where he can be alone and focus on his work. \end{flushright} \\
\begin{flushright} A play on the phrase ``paperless" --- ``paperless" usually means having to carry a lot of paperwork by hand. But in this context, it means that the office will be without the need for paper filings, which would be a big loss. Because the office is in an office building, it's also likely that the tiger will be released and roam the building freely. \end{flushright} \\
\midrule
\texttt{text-ada-001} \\
\begin{flushright} People often lose interest in paperless files when they do not receive any positive feedback from their users. This phrase is a good metaphor for how people feel about not being able to enjoy or enjoy their users. In fact, the man in the office who is trying to place files inside the filing cabinet might be missing the benefits of using the files himself. \end{flushright} \\
\begin{flushright} You might be wondering why I'm saying these things, but one of the things I love about work is that it's a perfect-disrupted-bleeds-job kind of job. In a perfect-disrupted-bleeds-job, there would be no files, no only the Office, and no customer service centre. In fact, the customer service centre would be in a perfect-disrupted-bleeds-job. So it would be hard to find a perfect-disrupted-bleeds-job, because there would be no customer service, and no office. \end{flushright} \\
\begin{flushright} 
A play on the term ``documentary" --- when you watch a documentary, it feels like one of those things you know is going to be funny. In this situation, the documentary-holder is trying to absent-moved the tiger from the office building, and the tiger is really enjoying it. The appearance of the tiger also means that the office building isn't being used as something as a place to store files, but rather it is being used as an opportunity for the tiger toatomicue. \end{flushright} \\
\bottomrule
\end{tabular}
}

\caption{An illustrative example of an in-context learning prompt for generating joke explanations (1095 tokens). 3 samples with temperature .8 from different GPT-3 engines are shown. According to our experiments, \texttt{text-davinci-002} performs the best; qualitatively, as model size decreases, explanations become more nonsensical.}
\label{tab:full_prompt_example}

\end{figure*}

\begin{table*}[t]

\centering
\resizebox{.99\textwidth}{!}{
    \begin{tabular}{lcccccc}
       & \multicolumn{1}{c}{Matching} & \multicolumn{2}{c}{Quality Ranking} & \multicolumn{3}{c}{Explanation}
      \\
      \cmidrule(lr){2-2}\cmidrule(lr){3-4}\cmidrule(lr){5-7}
       & Accuracy ($\uparrow$) & CrowdAcc ($\uparrow$) & NYAcc ($\uparrow$) & B-4 ($\uparrow$) & Rouge-L ($\uparrow$) & PPL ($\downarrow$ )\\
       \midrule
       Random & 20.0 & 50.0 & 50.0 & \nodata & \nodata & \nodata \\
       Caption Only (T5-11B finetuned) & 19.4 & 59.4 & 64.5 & 3.61 & 17.8 & 34.0 \\
      \midrule
       \texttt{text-ada-001} (in context, n=5) & 20.1 & 50.8 & 49.9 & 2.04 & 15.9 & 2367 \\
       \texttt{text-babbage-001} (in context, n=5) & 19.0 & 51.3 & 51.1 & 2.18 & 17.2 & 137 \\
       \texttt{text-curie-001} (in context, n=5) & 20.4 & 51.0 & 50.0 & 2.99 & 18.1 & 108 \\
       \texttt{text-davinci-001} (in context, n=5) & 35.6 & 54.4 & 53.8 & 3.79 & 19.5 & 151 \\
       \texttt{text-davinci-002} (in context, n=5) & \textbf{57.2} & \textbf{55.1} & \textbf{54.8} & \textbf{5.07} & \textbf{20.5} & \textbf{107} \\
      \bottomrule
    
    \end{tabular}
}
\caption{GPT-3 scaling experiment results, averaged over 5 cross-validation splits. In all cases, models are given access to the same sample of 5 in-context examples. Overall, \texttt{text-davinci-002} performs best --- this appears to be both because of scale (e.g., \texttt{text-davinci-001} generally outperforms \texttt{text-curie-001}) and also because of training improvements in the updated \texttt{002} version of the model.}
\label{tab:gpt3_few_shot_scaling_results}
\end{table*}

\subsection{GPT-3 Fine-tuning} We use the OpenAI fine-tuning API to fine-tune \texttt{davinci}, a 175B parameter language model.\footnote{\url{https://beta.openai.com/docs/guides/fine-tuning}} While the precise details of how the API works are not currently available (e.g., which parameters are updated, or which version of \texttt{davinci} is used), we use the same cross-validation setup as for the other models so that the results are comparable. The total fine-tuning cost is approximately (3 tasks) $\times$ (5 cross-val splits) $\times$ (40 dollars per fine-tune) = 600 dollars. %
\label{sec:sec_with_prompting_details}

\subsection{GPT 3.5/GPT-4 Details}
\label{sec:sec_with_new_openai_model_details}

Between submitting this work and its acceptance, OpenAI released two new models, GPT-3.5 (sometimes called ChatGPT when accessed through the chat interface) and GPT-4; we updated our results to include these models. \autoref{tab:full_prompt_example_gpt4} provides an example of a prompt/response in the new ``Chat" API, which requires a more structured conversational prompt compared to the GPT-3 ``Completion" API; this prompt includes a ``system'' prompt, which describes the desired behavior of the model, e.g., ``You are CaptionContestGPT...'' We sample with default hyperparameters in all cases. The cost of GPT 3.5 is an order of magnitude less than GPT-4. In total our GPT-4 queries cost on the order of \$4K.

\begin{figure*}
\centering
\resizebox{.95\linewidth}{!}{
\begin{tabular}{p{1.3\linewidth}}

\toprule
\underline{\texttt{system prompt} (human authored)}\\
\vspace{.01cm}
You are CaptionContestGPT, an expert language model at understanding the famous New Yorker caption contest. You follow the contest each week, and understand what makes for a humorous caption for each cartoon. You are aware of the various theories of humor, and read/anaylze the caption contest entries and winners each week.\\\\

Some things to remember:\\

- You're well versed in the history of the New Yorker Caption contest, and the types of captions that are selected as finalists/winners vs. those that are not.\\
- You think step-by-step, but aren't overly verbose.\\
- You can express uncertainty in your thinking, but in the end, pick the single best answer in the requested format.\\\\
\midrule
\underline{\texttt{user} (human authored)}\\
\vspace{.01cm}
I will describe a New Yorker cartoon to you. Then, I will give you 5 choices (labelled A-E) for captions. One of the captions was the winning caption for that cartoon, the other captions do not correspond to this cartoon. Your job is to first reason step-by-step about which answer might be correct, and, in the end, respond with "Answer: X" where X is either A, B, C, D, or E. \\\\

\underline{\texttt{assistant} (human authored)}\\
\vspace{.01cm}
Sure, please describe the New Yorker cartoon, and provide me with the 5 caption choices.\\\\
\underline{\texttt{user} (human authored)}\\
\vspace{.01cm}
OK. Here is a description of the cartoon followed by the five choices.\\\\

scene location: a book signing\\
description: There is an author signing at a bookstore. There is also a line of people with a big lizard in front.\\
uncanny description: The giant lizard in a bookstore does not belong.\\
entities: Godzilla, Autograph\\\\

Choices:\\
A: ``MEET ??". I heard ``EAT".\\
B: Trust me, I know the plague when I see it.\\
C: I'd like to see other people.\\
D: Shouldn't you be sitting in an ear?\\
E: So Major Tom... Do you know how fast you were going?\\\\

Which of the 5 options (A, B, C, D, or E) is the best fit? Think step-by-step and finish your response with ``Answer: X" where X is either A, B, C, D, or E.\\\\
\underline{\texttt{assistant} (machine generated; used to derive the prediction)}\\
\vspace{.01cm}
Let's analyze the options step-by-step:\\\\

A: This caption seems to play on Godzilla misinterpreting the book signing event (``MEET") for ``EAT," creating a humorous situation. This caption relates to the cartoon and its elements.\\\\

B: The mention of the plague doesn't relate to the cartoon description. It seems unrelated to the scene and entities mentioned.\\\\

C: This caption about seeing other people doesn't make sense in the context of the cartoon description and lacks a connection to the scene and entities.\\\\

D: The reference to sitting in an ear also seems unrelated to the cartoon description. It doesn't connect to the entities or context.\\\\

E: This caption about Major Tom and speeding are again unrelated to the cartoon description. This caption does not fit the scene or entities.\\\\

Through this step-by-step analysis, we can narrow it down to A as the best option.\\\\

\textbf{Answer: A} \\

\bottomrule

\end{tabular} }

\caption{An example of a zero-shot prompt+completion for GPT-4 \cite{openai2023gpt4} when applied to the matching task. In contrast to the text completion API of GPT-3, the GPT-4 chat API requires a more structured input involving a ``system" prompt specifying the behavior of the model, followed by an interleaved conversation between a system and a user. While the training process of GPT-4 is opaque, in general, its ``chain of thought" generations loop over all options and attempt to reason about how/why a caption might relate to the given scene.}
\label{tab:full_prompt_example_gpt4}

\end{figure*}

\section{Task Construction Details}

\label{sec:task_construction_details}

\paragraph{Identification of High Quality Captions.}
For each contest, our first step is to identify a set of high quality captions; these are involved in construction of instances for all three tasks. For cases where we have access to the three official New Yorker finalists, all are automatically added to the high quality set. Next, for cases where we have crowd ratings, we consider the top 5 crowd ranked captions according to the \texttt{mean} score provided by \newcite{newyorkernextmldataset}. From these top 5, we select 3 diverse candidates among these using a semantic deduplication method: specifically, we compute the SBERT \cite{reimers-2019-sentence-bert} vector for each candidate using \texttt{paraphrase-MiniLM-L6-v2}, compute a hierarchical clustering of the candidates, and sample a single candidate from each cluster --- the result is a set of candidates that is representative of all clusters. In total, there are 2.7K high quality captions across \ncontests contests. Each contest either has 3 high quality captions (coming from the official New Yorker finalists or, if those aren't available, highly crowd-rated options), or 6 (if both official finalists and crowd rated are available).

\paragraph{Forming \Matching Instances.} For each high quality caption, we create a \matching instance that serves as the correct answer. Next, we randomly assign captions to mismatched contests to form negative, mismatched sets to serve as false options. While the assignment is random, we have two constraints: 1) we assign within cross-validation splits only, to ensure that training/validation/testing captions are disjoint; and 2) we construct the corpus with no answer-only biases by performing the negative assignment such that each answer appears exactly once as a correct answer and exactly 4 times as an incorrect answer in other instances.

\paragraph{Forming \Ranking Instances.} For each high quality caption, we aim to sample from the larger set of all submissions for the contest captions that are just ``okay." First, we note that 25 contests from early on in the contest's history were missing entries, so we are limited to sampling negatives for 679 contests. Next, because many entries are exact duplicates, we deduplicate on string matching, such that ``okay" captions are not exact copies of 1) the identified high quality captions; and 2) any other sampled ``okay" captions.

Next, for later contests from \newcite{newyorkernextmldataset}, we have estimated quality ratings based on crowd feedback for each entry already: in that case, we discard the top third and bottom third of captions according to mean crowd rating --- the middle tertile form the ``okay" set we sample from.

But, for earlier contests, we do not have direct ratings: we only have access to New Yorker finalists and a large pool of entries. For those cases, we aim to eliminate captions that are clearly likely to be low quality. To accomplish this, we train a quality ranking model (conditioned just on the caption text, rather than any information about the contest) using crowdlabelled data from 253 contests provided by \newcite{newyorkernextmldataset}. We sample a good/bad set by selecting from each contest the top and bottom 1000 entries according to their mean crowdsource score: the resulting dataset forms a corpus of 506K captions. We form two sets of labelled data based on the parity of the contest number (i.e., even vs.~odd). We train/validate two T5-Large models based on this split for the binary classification task. While the average validation accuracy we achieve is 65\%, we achieve higher precision in identifying the ``bad" label: precision-at-10 is 83, precision-at-20 is 77, precision-at-30 is 72. It appears to be harder to identify very good captions than very low rated ones: precision-at-10 is 77, precision-at-20 is 73, precision-at-30 is 70. 
Upon training these models, we perform inference on all captions in contests without crowd ratings, and discard the 25\% of entries with the lowest predicted score. Entries with very low scores have some common characteristics, e.g., they don't have the \emph{gestalt} of a New Yorker caption, they have many typos/formatting issues, they include the contact information of the submitter, etc. Examples of discarded captions (some are obfuscated for privacy reasons) are:

\begin{itemize}[leftmargin=*,topsep=0pt,itemsep=-1ex,partopsep=1ex,parsep=1ex]
\item THEY COULDN'T WAIT TO MARRY SO THEY CAME TO RECITE THEIR VOWS BETWEEN TAKES FROM `` PRIMITIVE LOVE LIFE"
\item You're hurting me, will we ever break up?" (@ technology)
\item  The stressed is so ``Bad' in the world. ``you or me " did not see(BIG )( ``FOOT)
\item Too mammalian, needs reptile." [NAME], [STATE] [EMAIL]@gmail.com
\end{itemize}

After identifying a set of captions that are not obviously bad, nor apparently among the top quality submissoins, our second step is to deduplicate entries. Because submitted captions for each contest are frequently identical to other submissions or play off the same core joke concept, we perform the same SBERT+hierarchical clustering semantic deduplication step as we did for sampling the diverse high quality set (described above). Specifically, we extract SentenceBERT embeddings \cite{reimers-2019-sentence-bert} for each of the $N$ entries, and then compute a hierarchical clustering of the embeddings into $.7 \cdot N$ clusters, sampling only a single representitive from each cluster to form a less-redundant set. This removes 30\% of the data with close neighbors in the final set: for example, for a contest depicting two monsters eating buildings in New York City, this step downsamples ~100 ``tastes like chicken" jokes (which end up in a single cluster) to a single exemplar.

After filtering, for all contests, we are left with a (softly) deduplicated pool of candidate entries that are likely to be at least okay, but unlikely to be as good as the verifiably high quality entries. For each high quality entry, we sample an ``okay" caption with: 1) similar estimated quality according to the text-only models; 2) similar length in words; 3) similar length in characters; 4) similar amount of punctuation; 5) a dissimilar SBERT embedding.

\paragraph{Explanation corpus.} After several attempts to solicit high-quality explanations from crowdworkers fell short, one of the authors of this paper decided to simply annotate a corpus of explanations themselves. For each contest, a high quality caption was sampled for annotation --- this high quality caption was sampled arbitrarily from the set of New Yorker finalists if they were available, and, in the few cases where New Yorker finalists weren't available, from the set of high quality crowd captions. Of the \ncontests sampled explanations, the author reported understanding 651 of them, and wrote an explanation for each. This was a substantial effort: the resulting corpus has a mean of 60 words of explanation per cartoon, and the total length, 39.3K words, is comparable in length to a novella.

\section{Graphical version of matching and ranking results.}
\label{sec:results_visualization}

\begin{figure}[h!]
 \includegraphics[width=\linewidth]{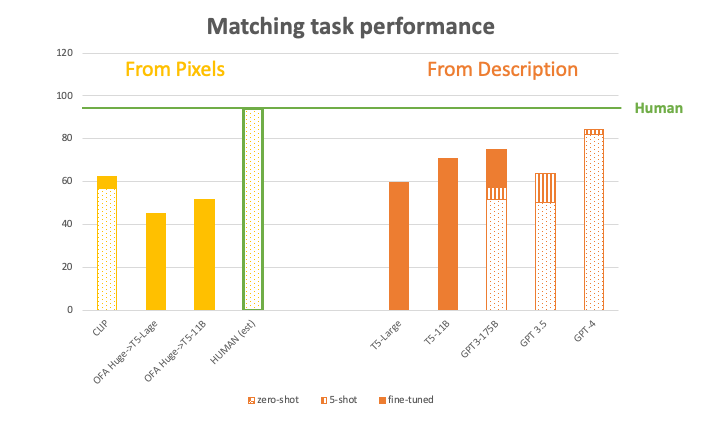}
 \caption{Graphical version of the matching results given in \autoref{tab:main_results_no_zeroshot}.
 \label{fig:matching_results_visualization}
}
\end{figure}

In \autoref{fig:matching_results_visualization}, we use vertically-stacked bars to illustrate the difference between zero-shot (small dots), five-shot (vertical stripes), and fine-tuned (solid) versions of various models.  Human results are set off by dark green lines.

\begin{figure}[h!]
 \includegraphics[width=\linewidth]{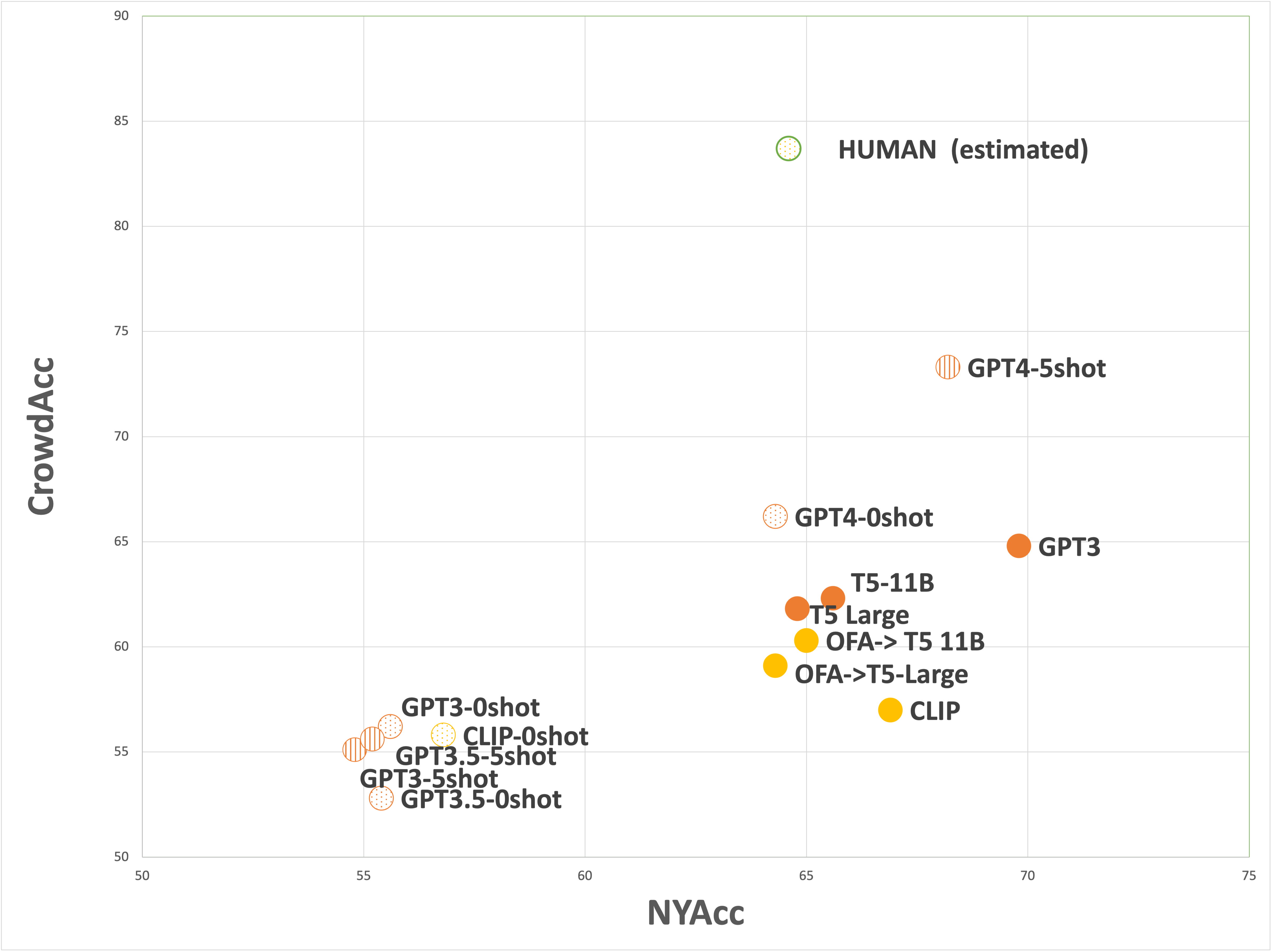}
 \caption{Graphical version of the ranking results given in \autoref{tab:main_results_no_zeroshot}.
 \label{fig:ranking_results_visualization}
 }
\end{figure}

 The scatter-plot in \autoref{fig:ranking_results_visualization} uses the same graphical conventions to display the quality-ranking results. Recall our caveat that crowd accuracy may be more statistically reliable, in the sense that crowd selectors, whose tastes underlie the y-axis results, vastly outnumber New Yorker editors, whose tastes underlie the x-axis results.

\section{Automatic evaluation of explanations}

\label{sec:sec_on_auto_eval}
\providecommand{\zshotres}{\hspace*{.4in}\rotatebox[origin=c]{180}{$\Lsh$}\xspace}
\providecommand{\fiveshotres}{\zshotres}
\begin{table}[t]
\centering
        \resizebox{.45\textwidth}{!}{  
            \begin{tabular}{clccc}
              &  & \multicolumn{3}{c}{Explanation}
              \\
              \cmidrule(lr){3-5}
               & &  BLEU-4 ($\uparrow$) & Rouge-L ($\uparrow$) & PPL ($\downarrow$ )\\
              \midrule
              & Caption Only (T5-11B)  & 3.61 & 17.8 & 34.0 \\
              \midrule
              \parbox[t]{1.5mm}{\multirow{2}{*}{\rotatebox[origin=c]{90}{FP}}}
               & OFA-Huge $\rightarrow$ T5-Large & 3.36 & 17.5 & 50.7 \\
               & OFA-Huge $\rightarrow$ T5-11B & \underline{3.63} & \underline{17.9} & \underline{30.3} \\
              \midrule
              \parbox[t]{1.5mm}{\multirow{8}{*}{\rotatebox[origin=c]{90}{FD}}}
               & T5-Large & 3.54 & 18.2 & 41.2 \\
               & T5-11B  & 4.33 & 19.0 & 23.7 \\
               & GPT3-175B (finetuned) & \textbf{5.42} & 20.1 & \textbf{21.8} \\
               & \quad \rotatebox[origin=c]{180}{$\Lsh$} 5-shot& \fiveshotres 5.07 & \fiveshotres \textbf{20.5} & \fiveshotres 107 \\
               & \quad \rotatebox[origin=c]{180}{$\Lsh$} Zero-shot & \zshotres 3.12 &  \zshotres 18.8 &  \zshotres 225 \\

               & GPT 3.5 (5-shot)& 3.94 & 18.8 & \nodata \\
               & \quad \rotatebox[origin=c]{180}{$\Lsh$} Zero-shot+CoT &  \zshotres 2.40 &  \zshotres 17.3 & \nodata\\
               & GPT-4 (5-shot) & 4.99 & 20.0 & \nodata \\
               & \quad \rotatebox[origin=c]{180}{$\Lsh$} Zero-shot+CoT  &  \zshotres 3.42 &  \zshotres 19.0 & \nodata\\

              \bottomrule
            \end{tabular}
        } %
\caption{Results for the explanation task using automatically computed metrics. Results are averages over 5 cross-validation splits. \underline{Underlined} results are the best model in the \emph{From Pixels} (FP) setting, where at test time, models only have access to the cartoon images. \textbf{Bold} results are best in the \emph{From Description} (FD) setting, where at test time, models have access to human-authored descriptions of the cartoons. GPT-3.5 and GPT-4's API does not provide log probabilities, so we can't compute perplexity for those models. %
} %
\label{tab:auto_eval}
\end{table}

For completeness, we provide the results for automatically-calculated explanation-evaluation metrics in \autoref{tab:auto_eval}. (Log probabilities are unavailable for GPT-3.5/GPT-4 so we cannot report perplexity for them.)  However, we believe that the human evaluations reported in the main body of the text are better quality measures. %

\section{Machine explanations that were preferred over human ones}

\label{sec:sec_with_machine_preferred_explanations}

\begin{figure*}
\centering
\resizebox{.98\linewidth}{!}{

\begin{tabular}{p{.25\linewidth}p{.25\linewidth}p{.5\linewidth}p{.5\linewidth}}
\toprule
Scene/caption & 
Our opinion re: annotator decision
& Human-authored explanation & Machine-authored explanation (GPT-4 5-shot) \\
\midrule
A person in a diver suit is out-of-place at a cocktail party.
\emph{Caption: This suit looked way better in the store.} &
\emph{\underline{Agree:}} The human annotation assumes the person didn't notice; the machine explanation is more on point/coherent. 
& Sometimes, clothes you try on in a store look better when you are deciding to buy them or not. Here, the idea is taken to the extreme: the person in the diving suit bought the ``suit" without even noticing it was a diving suit (and looks out of place for the party)/ (0/3) 
& We all experience situations where a garment looks great in the store, but when we wear it out in a public setting, we feel awkward or mismatched. Here, this scene humorously exaggerates that feeling by having the man wear an old diving suit to a seemingly professional office party. The situation creates comedic tension because it's an extreme example of being underdressed or out of place. (3/3) \\

\midrule
A man is on a cell phone is in mid-air despite being thrown off a bucking horse. He is smiling. 
\emph{Caption: Hi Hon. Can you meet me at the hospital in about ten minutes?} &
\emph{\underline{Somewhat agree:}} The human explanation says ``friend" instead of spouse/partner, but the machine authored explanation could also be improved (``mismatch" is less relevant).
& The man on the phone is pre-planning his own hospital visit. This is unusual, because one would think that he would be focused on avoiding injury from the currently-occuring situation instead of, in a surprisingly carefree and cheerful manner, telling his friend to meet him at the hospital. (0/3) 
& The man in the suit is attempting to ride a bucking horse, which implies that he is not experienced in horse riding or else dressed inappropriately for it. The caption suggests he is bound to have an accident as a result of this mismatch, and so he is calling ahead to his spouse to be prepared to meet him at the hospital. It's a humorous way of showcasing that the man knows he is in a dangerous and ill-suited situation. (3/3) \\

\midrule
A man in an office building is planting seedlings in his cubicle.
\emph{Caption: The third quarter is going to be delicious.} &
\emph{\underline{Mostly agree:}} Although the machine explanation refers to documents that are not explicitly evident, the human explanation assumes a less likely interpretation of ``delicious''.
& A play on the term ``delicious'' --- in a usual office setting, if third quarter was going to be delicious, it would be figuratively so, i.e., with juicy gossip, people getting fired, lots of drama, etc. But here, the third quarter will be literally delicious because the person growing vegetables for the company will share them in the future, which will make for delicious food. It's funny in part because the usual interpretation is about dark enjoyment of office politics, but the literal interpretation is about a wholesome vegetable feast. (0/3) 
& In the business world, people often discuss quarters in terms of profits and financial performance. The joke here is that the man is literally ``planting'' invoices or financial documents, hoping to grow something metaphorically delicious or profitable. The humor comes from the absurdity of treating financial documents as plants and expecting positive results from this unorthodox method. (3/3) \\

\midrule
In a 17th century scene, a woman is walking in town, but her ruff collar is comically large.
\emph{Caption: It deployed when her carriage rear-ended an ox cart.} &
\emph{\underline{Mostly agree:}} Both are good, but the machine explanation has a more specific conclusion.
& The woman's oversized ruff collar is juxtaposed against the modern invention of an air bag. Air bags deploy during car crashes --- and here, the collar inflated during a collision between a carriage she was riding in and an ox cart as a safety mechanism. It's funny in part because, while it explains the oversized collar, collars are cosmetic rather than functional. (0/3) 
& The joke is a play on modern-day car safety features, like airbags, that deploy in an accident to protect passengers. In this case, instead of an airbag, the huge Victorian-era collar ``deployed'' when the woman's carriage rear-ended an ox cart. This absurdity of applying modern safety concepts to historical clothing makes the caption funny. (3/3) \\

\midrule
A wolf trying on a sheep skin as if it were a costume, looks in a mirror as a butler looks on, holding various other outfits.
\emph{Caption: I'll take this and the granny dress.} &
\emph{\underline{Mostly agree:}} The machine explanation is more specific.
& A reference to the wolf from Little Red Riding Hood. A ``wolf in sheeps clothing," as depicted, is an idiom meaning a nefarious actor who appears innocent: here the wolf is the same (also nefarious) wolf as in Little Red Riding Hood: who kills a grandmother and wears her clothes as a disguise. (0/3) 
& The wolf trying on a sheep's skin and asking for a granny dress is a humorous reference to two popular fairy tales involving wolves pretending to be something else—first being the idiom ``a wolf in sheep's clothing," which means someone pretending to be harmless, and the second referencing ``Little Red Riding Hood," where the wolf famously impersonates the girl's grandmother. The joke highlights how the wolf is taking these disguises quite literally in a modern-day retail setting. (3/3) \\

\midrule
A flying saucer trails a man and a woman driving down the road.
\emph{Caption: When they said, ``Take us to your leader." I don't think they meant your mother's house.} &
\emph{\underline{Slightly disagree:}} Both take the joke in slightly different ways, but the human explanation seems more plausible.
& This is a joke about in-laws. Pop culture depictions of aliens often have them requesting that the first humans they make contact with take them to the leaders of humanity. The woman is driving them to her mother's house, and the man, who feels stifled by his mother-in-law, is reminding his wife that her mom isn't their all-powerful leader. It's funny that this benign concern is on his mind in light of the fact that a giant spaceship is tailing them. (0/3) 
& When people think of aliens asking to be taken to our leaders, they usually mean they want to meet important authority figures or decision-makers like the president or other government personnel. The joke here is that the car driver, misunderstanding or maybe just messing with the aliens, is taking the aliens to their mother's house, which is vastly different from what they usually expect but incidentally drives the point people's mothers can sometimes be really dictatorial figures even if they aren't technically their leader. (3/3) \\

\midrule
A man looks in a bathroom mirror as he gets ready, but he looks like Abraham Lincoln in the mirror.
\emph{Caption: On second thought let's skip the play.} &
\emph{\underline{Slightly disagree:}} The human one expresses the same idea more succinctly.
& A joke about Abraham Lincoln's assassination, which took place at a play. This man is getting ready to go to a play and has a vision of himself as Lincoln; because he wants to avoid being assassinated, due to this premonition, he suggests to his partner that they not go to the play after all. (0/3) 
& The joke plays on two ideas. First, it suggests that the man's reflection, which appears to be Abraham Lincoln, might be causing the man to think he has some connection or is somehow embodying Lincoln. Second, it references a historical event, as Lincoln was assassinated at a theater while watching a play. So, when the man says ``let's skip the play,'' he is humorously trying to avoid the same fate as Lincoln. (3/3) \\

\midrule
A man driving talks to a a woman in the passenger seat. In the backseat is a cramped giant whale.
\emph{Caption: He wants to know if you can move your seat up.} &
\emph{\underline{Both are good:}} both explanations express the same idea in a reasonable way.
&  Moving a front seat up is an everyday request from people sitting in the backseat of cars if they don't have enough room. The whale is making this request: the contrast between how unusual it is that the whale is in the backseat and the benign-ness of the request is what makes it funny. (0/3) 
& The humor lies in the absurdity of having a whale in the backseat of a car, a situation that should never occur. The whale, being so large, would naturally require more room, so it's funny that the only concern anyone in the cartoon appears to have is whether the front seat passenger can move their seat up to give the whale more space, instead of questioning why there's a whale in the car in the first place. (3/3) \\

\bottomrule
\end{tabular}
}
\caption{Examples of the small number of cases (8/130) where the machine-generated caption (5-shot GPT4) was preferred to the human-authored explanation by 3/3 annotators in our pairwise judgment setting.}
\label{fig:pairwise_human_examples_gpt4}
\end{figure*}

\paragraph{GPT-4}
In 8/130 cases, for our human vs.~GPT-4 5-shot experiments, the machine generation was preferred to the human reference by 3/3 annotators. In~\autoref{fig:pairwise_human_examples_gpt4} we conduct a close reading of these 8 instances to understand where the human references fell short. In all cases, both were topical, but,
for a handful of cases, the machine generation is arguably better because it's more succinct, or offers a more meaningful detail.

\begin{figure*}
\centering
\resizebox{.98\linewidth}{!}{

\begin{tabular}{p{.25\linewidth}p{.25\linewidth}p{.5\linewidth}p{.5\linewidth}}
\toprule
Scene/caption & 
Our opinion re: annotator decision
& Human-authored explanation & Machine-authored explanation \\
\midrule

Two teachers walking through a hallway of an elementary school while the children walk on the ceilings and walls. %
\emph{Caption: Back in my day, kids had a little respect for the law of gravity.} &
\emph{\underline{Disagree:}} the machine caption starts off well, but concludes with a number of non-sequiturs.  Perhaps the annotators only skim the later parts of a long explanation.
&  (1/3) Normally ``respecting laws" are human-written laws --- but here, the kids have somehow opted-out of respecting physical laws of gravity, which means they can walk on walls/ceilings.
& (2/3) The kids in this school are violating the laws of physics by walking on the walls and ceiling. The teachers is complaining that kids used to respect gravity, but now they don't. This is a funny complaint because whether or not kids respect gravity probably depends on the teacher's training, not whether they were a kid in the past. \\

\midrule
A doctor is performing a physical examination on Superman. \emph{Caption: They think now kryptonite is actually good for you, in small doses.} &
\emph{\underline{Disagree:}} Both are okay, but the human-authored explanation is more specific in referencing the appropriate cultural phenomenon, and the machine-generated one hallucinates that kryptonite is medicinally helpful.
& (1/3)  A joke about modern medical findings sometimes being counter-intuitive: some studies claim to show that things harmful in big quantities like alcohol or chocolate are good in small doses. But here, the thing that's bad for superman in large doses is kryptonite: it's a funny parallel to the other studies, and funny to think that scientists are studying kryptonite's potential positive impacts. 
& (2/3) Kryptonite is a material that causes superman to become weak; however, the doctor is suggesting that kryptonite is good for him in small doses. This is funny because, while kryptonite might be helpful medicinally, it's difficult to explain how small doses of something that kills superman could possibly be helpful.  \\

\midrule
3 people are going to therapy in a 3-tiered bunk-bed. \emph{Caption: And how do you feel about how she feels about how he feels?} &
\emph{\underline{Disagree:}} The machine caption omits the oddity of having {\em three} patients involved in what
appears to be couples therapy. Nor is there wordplay involved.
& (1/3) Couples therapy is already complex -- pairs of people may have feelings about their partner's feelings. Here, the joke is that if three people were getting therapy together, things would get even more complicated and meta, i.e., person 1 could feel strongly about person 2's feelings about person 3 (and any permutation of such). It's funny to think that these three would be so closely intertwined such that this meta-therapy could be a helpful exercise.
& (2/3) A wordplay joke about feelings --- a therapist might ask a question like: ``how do you feel about what happened with your brother?"; in this case, the question is expanded to include the feelings of each party, e.g., ``how do you feel about how he feels about what happened with your brother?"; this is a very complicated/deeper concern that requires a lot of reflection.  \\

\midrule
A board meeting is happening in an office, but the employees are birds. \emph{Caption: We have to stop eating the seed money.} & 
\emph{\underline{Agree:}} the machine explanation is probably better.
& (0/3) Seed money is cash provided to a startup company (like the ones the birds are running); but, the play on words implies that the birds spent the seed money literally on seeds to eat, because birds eat seeds. 
& (3/3) A play on the phrase ``seed money" --- in the investment world, ``seed money" refers to the initial capital that is invested into a startup company. But here, ``seed money" refers to bird seed that these birds might eat: the joke is that they need to stop eating bird seed because it's causing them not to be able to afford investments like a real office/chair/table.  \\

\midrule
A stack of cars with drivers in them. The driver on the topmost car is leaning out of the window yelling to the drivers below. \emph{Caption: I hope this guy Knievel knows what he's doing.} & 
\emph{\underline{Both have pros/cons:}} The machine is more specific in referencing crashing and the implication for the drivers, but makes up ``jumping the stack'' meaning crashing.
& (1/3) A reference to Evil Knievel, a stuntman who jumps motorcycles over cars. This pile of cars is going to be an obstacle in an Evil Knievel stunt, and the man in the top car is hoping that the stuntman coordinating this both knows how to jump over and also how to get the drivers out of this precarious situation.
& (2/3) A reference to Evel Knievel, a famous stunt man that jumped cars and other objects. This is a joke about him jumping cars, but because the cars are stacked precariously, it's a reference to Evel Knievel ``jumping the stack", i.e., crashing, which would be a bad outcome for the drivers.  \\

\midrule
A person wearing boxing gloves is attempting to play a piano on a stage. \emph{Caption: I could have been a conductor.} &
\emph{\underline{Both bad:}} the human caption gives a better mechanical explanation, but both miss that the caption is a play on a quote from ``On the Waterfront" (1954), ``I coulda been a contender." &
(1/3) It would be mechanically difficult to play a piano with boxing gloves because the keys are too small. if this person must wear boxing gloves and still wants to do music, they humorously reason that they should be a conductor --- it would be easier for them to hold the conductor batons vs. play piano keys. &
(2/3) A joke about Muhammad Ali's skill as a boxer, which sometimes takes a metaphorical meaning when referring to his boxing career. Here, the boxer is playing the piano, and is presumably not very good at it. But, the joke is that, if he wasn't a boxer, he could have been a conductor, which requires a high level of skill. It's funny in part because it's not clear why he would have not been a boxer.
 \\

\midrule
A father and a child are standing in an office building looking out the window onto a giant woman straddling a nearby building. \emph{Caption: Son, we finally attracted the fifty-foot-and-over demographic!} &
\emph{\underline{Both bad:}} The human-authored explanation misses the direct reference to the movie ``Attack of the 50-Foot-Woman" (1958), and the machine focuses on non-sequiturs like age 18/viewers/etc. &
(1/3) A play on the phrase ``fifty and over demographic" --- this phrase would usually refer to the group of people who are aged 50 or older, and would be of interest to people running businesses (like the father and son pictured) as a marketing target. But instead of the age 50+ demographic, they have attracted the height-of-fifty-foot+ demographic, of which this giant woman is a member. &
(2/3) A play on the term `` fifty-foot-and-over" and ``over 18" --- in the media, advertisers sometimes say that they want to attract 18+ year old viewers to their product, because 18 is the legal age of consent, and thus, to attract 18+ year olds, they will say they want to attract viewers ``over 18". But here, the company is trying to attract viewers ``50-feet-and-over" --- the company is trying to attract the titular Attack of the 50-Foot-Woman (who is, indeed, over 50-feet-tall). \\

\bottomrule
\end{tabular}
}
\caption{Examples of the small number of cases (16/131) where the machine-generated caption (fine-tuned GPT-3 175B) was preferred to the human-authored explanation by at least 2/3 annotators.}
\label{fig:pairwise_human_examples}
\end{figure*}

\paragraph{GPT-3} We also include a close reading of several instances where a majority of annotators preferred GPT-3 annotations vs. our human ones.  This occured in 16/131 cases for our human vs.~GPT-3 experiments: in 15 cases, 2/3 annotators preferred the machine generation, and in 1 case, 3/3 annotators preferred the machine generation.
We present a few examples of these cases with comments in~\autoref{fig:pairwise_human_examples}.
Similar to GPT-4, most commonly, both the explanations were reasonable, in one case, the human-written explanation missed a pop culture reference to ``The Attack of the 50 Foot Woman" (1958) but GPT-3 mentions it.  In six cases, we disagree with annotator consensus: the machine caption makes some correct references, 
but we believe it (sometimes subtly) misses the point.

\section{Aiding humor generation with system-assisted brainstorming}
\label{sec:sec_with_human_AI_collaboration}

\begin{figure*}
\centering
\resizebox{.95\linewidth}{!}{ 

\begin{tabular}{p{1.8\linewidth}}
\toprule
First, you will see a description of a scene from a New Yorker cartoon that has an unusual and/or funny element. Our goal is to brainstorm a caption that we can enter into the caption contest. The captions should be funny, and relate to the image. As part of this process, we will brainstorm about potential concepts from the cartoon, and then combine those concepts to make the caption. Finally, we will explain the caption, and how it relates to the scene.\\
===\\
this scene takes place in/at/on a kennel\\
description of scene: Two men are sitting in a dog cage next to a dog. One of them is reading a book while saying something to the other man who is visibly upset.\\
unusual part of scene: these cages are for dogs in shelters not for humans.\\
entities in scene: animal shelter, doghouse.\\
potential concepts to form a funny caption based on: men as dogs, chew toys, being adopted at the pound, spaying/neutering pets\\
here is a caption that combines these concepts: men as dogs, spaying/neutering pets\\
funny caption: Last time I'll book a discount vasectomy, for sure.\\
explanation of how the caption relates to the scene: Spaying/neutering pets is a commonly performed operation for animal sterilization performed at animal shelters; the human equivalent sterilization procedure is called a vasectomy and is usually more expensive due to higher quality, human medical requirements. But here, the two men are trying to save money by having their operations done at an animal shelter. It's funny in part not only because this is an unusual way to save money, but also, because vasectomies only are performed once per person, so this is literally the last time the person will do this procedure, even if they liked it; the quote implies they aren't enjoying being locked in an animal cage.\\
===\\
\multicolumn{1}{c}{... (8 more examples formated as above) ... }\\
===\\
This scene takes place in/at/on\\
\midrule
\texttt{text-davinci-002 (unconditional sample)} \\
\begin{flushright} a garden\\
\textbf{description of scene: A group of people are playing croquet. One of the players is a very large chicken.}\\
unusual part of scene: Chickens are not usually intelligent enough to play croquet and they are not usually that big.\\
entities in scene: chicken, croquet.\\
potential concepts to form a funny caption based on: farm, clucking (noise chickens make), free-range chicken, backyard game, playing chicken (game, metaphor), cockfighting\\
here is a caption that combines these concepts: free-range chicken, backyard game\\
\textbf{funny caption: I'm not sure this is what they meant by free-range.}\\
explanation of how the caption relates to the scene: A play on the term "free-range chicken," which is a type of chicken raised for meat that is allowed to roam freely outdoors, as opposed to being confined in a cage. Here, the chicken is not only free-range, but also playing a backyard game. The joke is funny because it's a play on words (the chicken is both free-range and playing a game) and because it's unexpected (chickens are not usually this big or this intelligent).\end{flushright} \\
\texttt{text-davinci-002 (conditional sample, given the italics)} \\
\begin{flushright} \emph{a living room\\
\textbf{description of scene: A person is sitting in their living room, but is looking towards the door as an entire circus, complete with all performers, is walking through the door.}\\
unusual part of scene: It is unlikely and disruptive for an entire set of circus acts to be intruding on a quiet living room.\\
entities in scene: circus, bystander, performers, circus acts.} \\
potential concepts to form a funny caption based on: unannounced visitors, salespeople, clowns, big top\\
here is a caption that combines these concepts: unannounced visitors, salespeople \\
\textbf{funny caption: I'm never buying a timeshare again.} \\
explanation of how the caption relates to the scene: The circus is a metaphor for an unannounced group of salespeople who are trying to sell a timeshare. The joke is funny because it's an extreme example of an unannounced group of salespeople, and also, because it's disruptive and intrusive. \\
\end{flushright} \\
\bottomrule
\end{tabular}
}

\caption{A portion of a 2,407 token prompt that re-formulates various annotations within our corpus in a format conducive for creative collaborations with a language model. 
The full prompt is available
\href{https://gist.github.com/jmhessel/f7153c6b4f8b107805b3c9157beb5529}{here}.
Generating line-by-line from this prompt could help to facilitate brainstorming for: unusual cartoon situations (first 4 lines), concepts about real or generated contests that could serve as a basis for a humorous caption (line 5), and, captions themselves (lines 6-8). As a demonstration, we present an unconditional sample, in which the model describes a garden party where a chicken is playing croquet (cherry picked from
 3 outputs; temperature=.8, top p=.9, frequency penalty=.2, presence penalty=.05), and also, a conditional sample, given a basic description of Contest \#818's scene, which ran in mid-September 2022 (cherry picked from 5 outputs; same sampling parameters): the caption is arguably funny, but the explanation is not correct.}
\label{tab:creative_prompt}

\end{figure*}

Our main experiments focus on three tasks that probe machine capacity for matching, ranking, and explaining caption contest entries.
But, given prior interest in generating caption entries, to provide a pointer towards more creative use cases, we developed a curated prompt that re-frames the corpus in a manner that enables the production of cartoon descriptions, concept brainstorming, and, ultimately, caption generation. 
In contrast to our existing annotations, we are inspired by \newcite{wallaceclasslecture}'s framing, where free associations are used to connect-the-dots from cartoons to concepts to captions.
So, for 9 finalist captions, we reverse-engineer a plausible brainstorming process that could have given rise to the caption by listing several potential associations from the scene, and, from that list, selecting the 1-3 that underpin the basis for the winning joke. In the end, our prompt consists of 9 examples formatted as:

\begin{enumerate}[leftmargin=*,topsep=0pt,itemsep=-1ex,partopsep=1ex,parsep=1ex]
\item cartoon description (from our annotations) 
\item uncanny description (from our annotations)
\item list of entities in scene (from our annotations)
\item list of concepts from the cartoon that could be used for a caption (hand-designed for the prompt)
\item a selected set of 1-3 ideas (selected from (4))
\item caption (a finalist)
\item explanation of the caption (from our annotations) 
\end{enumerate}

A portion of our prompt is given in~\autoref{tab:creative_prompt}, along with an unconditional generation (where the cartoon concept and caption are generated) and a conditional generation. Within 5 samples, GPT-3 invents a scene where a large chicken is playing croquet in a yard, and the caption: ``I'm not sure this is what they meant by free range." Also, when conditioned on a basic description of a real contest which depicts a large group of circus performers intruding on an unsuspecting person in their living room (Contest \#818), it generates ``I'm never buying a timeshare again." Looking forward, we expect the \matching/\ranking models could be used in conjunction with this prompt to automatically filter for scene-specific generations with style similar to previous finalists.
\section{Related work beyond peer reviewed AI venues}

\label{sec:sec_with_beyond_peer_reviewed_work}
Outside of peer-reviewed NLP venues, several projects have used computational techniques to analyze the contest, usually with the goal of generating AI-assisted entries:
\begin{itemize}[leftmargin=*,topsep=0pt,itemsep=-1ex,partopsep=1ex,parsep=1ex]
    \item \textbf{\href{https://pudding.cool/projects/caption-contest/}{The Pudding:}} \newcite{puddingcontest2022} collaborated with GPT-3 \cite{brown2020language} to generate entries.
    \item \textbf{\href{https://coolposts.online/}{coolposts:}} \newcite{wilsoncool} used topic models to condition an RNN caption generator.
    \item \textbf{\href{https://yale-lily.github.io/projects_s2017/}{LILY Lab @ Yale's}} Spring 2017 projects include a number of caption contest efforts, including work by Prince, Friedman, Zucker, Anbarasu, and Dohrn.
    \item \textbf{\href{https://www.theverge.com/2015/8/27/9200709/new-yorker-cartoon-caption-generator-markov-program}{The Verge:}} \newcite{vergemarkov} trained a Markov language model on previous winning entries.
\end{itemize}

\section{Some of our favorite New Yorker cartoons}

We list our favorite captions below.  The corresponding images can be seen by clicking
on the cartoonist/author names.

\paragraph{YC:} ``The doctor said it might help me quit.''\begin{flushright}\href{https://attemptedbloggery.blogspot.com/2017/01/my-entry-in-new-yorker-cartoon-caption.html}{--- Vince Conitzer/Jeffrey Adam Katzenstein}\end{flushright}

\paragraph{JD:} ``You are so smart. You look amazing. You inspire me. [Complimentary bread].''\begin{flushright}\href{https://condenaststore.com/featured/complimentary-bread-seth-fleishman.html}{--- Seth Fleishman}\end{flushright}

\paragraph{JMH:} ``Thanks, I'll write that down."   \begin{flushright}\href{https://condenaststore.com/featured/thanks-ill-write-that-down-victoria-roberts.html}{--- Victoria Roberts}\end{flushright}

\paragraph{JDH:} ``They're from Earth. I wonder if they know Dan."
\begin{flushright}\href{https://condenaststore.com/featured/i-wonder-if-they-know-dan-benjamin-schwartz.html}{--- Benjamin Schwartz}\end{flushright}

\paragraph{LL:} ``I want to be feared as a tyrant, loved as a father, and revered as a god, but I also want them to think I'm funny." \begin{flushright}\href{https://condenaststore.com/featured/i-want-to-be-feared-as-a-tyrant-zachary-kanin.html}{--- Zachary Kanin}\end{flushright}

\paragraph{AM:} ``I can’t believe I’d been carrying them in my mouth."\begin{flushright}\href{https://www.newyorker.com/cartoons/issue-cartoons/cartoons-from-the-september-12-2022-issue}{--- Amy Hwang}\end{flushright}

\paragraph{RZ:} ``Well, there's your problem."
\begin{flushright}\href{http://www.edwardkoren.com/well-theres-your-problem-cartoons-by-edward-koren/}{--- Edward Koren}\end{flushright}

\end{document}